\documentclass[runningheads]{llncs}

\usepackage{eccv}

\usepackage{eccvabbrv}

\usepackage{graphicx}
\usepackage{booktabs}

\usepackage{colortbl}  %彩色表格需要加载的宏包
\usepackage{xcolor}
\usepackage{amsmath}
\usepackage{amssymb}
\usepackage{color}
\usepackage{multirow}
\usepackage{indentfirst}
\usepackage{threeparttable}
\usepackage{pifont}
\usepackage{makecell}
\usepackage{arydshln}

\usepackage[accsupp]{axessibility}  % Improves PDF readability for those with disabilities.

\usepackage[pagebackref,breaklinks,colorlinks,citecolor=eccvblue]{hyperref}
\usepackage{orcidlink}
\definecolor{myred}{rgb}{.99,.90,.90}

\begin{document}

\title{Exploring the Naturalness of AI-Generated Images} 


%\author{Zijian Chen\inst{1}\orcidlink{0000-1111-2222-3333} \and
%Second Author\inst{2,3}\orcidlink{1111-2222-3333-4444} \and
%Third Author\inst{3}\orcidlink{2222--3333-4444-5555}}
\author{Zijian Chen\inst{1} \and
Wei Sun\inst{1} \and
Haoning Wu\inst{2} \and Zicheng Zhang\inst{1} \and \\Jun Jia\inst{1} \and Zhongpeng Ji\inst{3} \and Fengyu Sun\inst{3} \and Shangling Jui\inst{4} \and Xiongkuo Min\inst{1} \and Guangtao Zhai\inst{1} \and Wenjun Zhang\inst{1}}

\authorrunning{Z.~Chen et al.}

\institute{Shanghai Jiao Tong University, China \and
Nanyang Technological University, Singapore \and
Huawei, China \and
Huawei Kirin Solution, China\\
\url{https://github.com/zijianchen98/AGIN}
}

\maketitle
\vspace{-0.5cm}

\begin{figure}
\setlength{\abovecaptionskip}{0.1cm}
\setlength{\belowcaptionskip}{-0.7cm}
  \centering
   \includegraphics[width=1\linewidth]{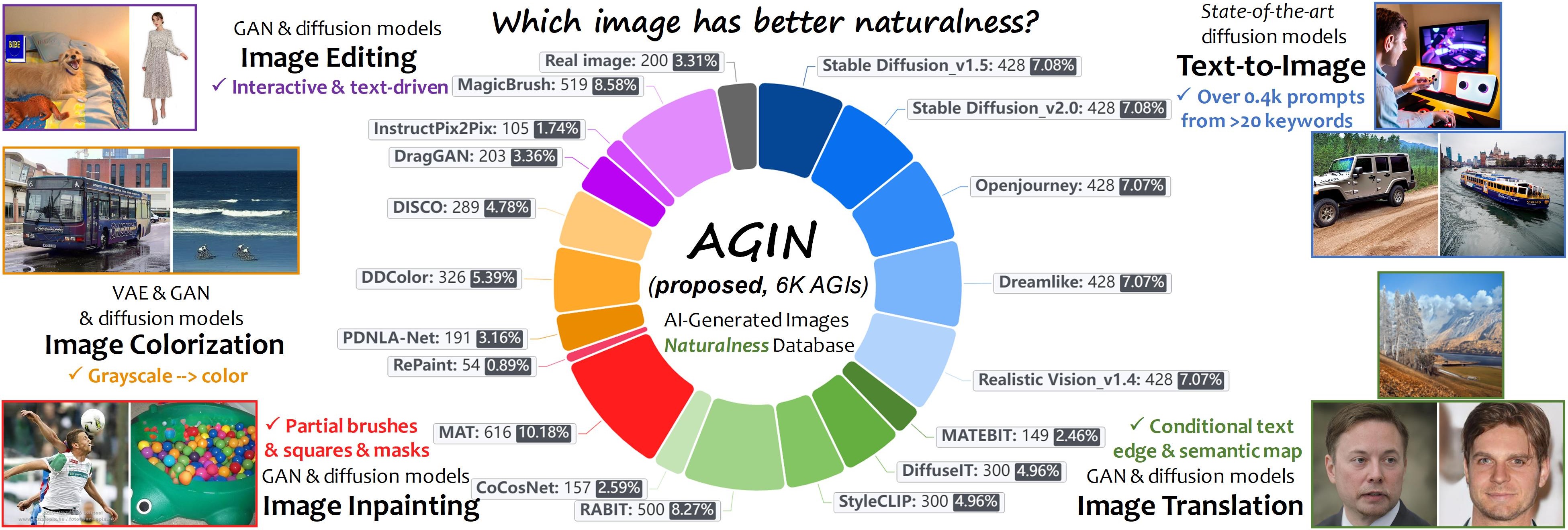}
   \caption{The proposed {\bf AGIN}, \textit{first-of-this-kind} image naturalness assessment database with human opinions from {\it technical}, {\it rationality}, and overall \textit{naturalness} perspectives,  focusing on 5 generative tasks ({\it i.e.}, {\it text-to-image}, {\it image translation}, {\it image inpainting}, {\it image colorization}, and {\it image editing}).}
   \label{intro}
\end{figure}

\begin{abstract}
The proliferation of Artificial Intelligence-Generated Images (AGIs) has greatly expanded the Image Naturalness Assessment (INA) problem. Different from early definitions that mainly focus on tone-mapped images with limited distortions (\eg, {\it exposure}, {\it contrast}, and {\it color reproduction}), INA on AI-generated images is especially challenging as it has more diverse contents and could be affected by factors from multiple perspectives, including low-level technical distortions and high-level rationality distortions. 
In this paper, we take the first step to benchmark and assess the visual naturalness of AI-generated images.
First, we construct the \underline{A}I-\underline{G}enerated \underline{I}mage \underline{N}aturalness (\textbf {AGIN}) database by conducting a large-scale subjective study to collect human opinions on the overall naturalness as well as perceptions from technical and rationality perspectives. AGIN verifies that naturalness is universally and disparately affected by technical and rationality distortions. 
Second, we propose the \underline{J}oint \underline{O}bjective \underline{I}mage \underline{N}aturalness evalua\underline{T}or (\textbf{JOINT}), to automatically predict the naturalness of AGIs that aligns human ratings. Specifically, JOINT imitates human reasoning in naturalness evaluation by jointly learning both technical and rationality features. We demonstrate that JOINT significantly outperforms baselines for providing more subjectively consistent results on naturalness assessment.
\end{abstract}

\begin{figure}
\setlength{\abovecaptionskip}{0.1cm}
\setlength{\belowcaptionskip}{-0.5cm}
  \centering
   \includegraphics[width=1\linewidth]{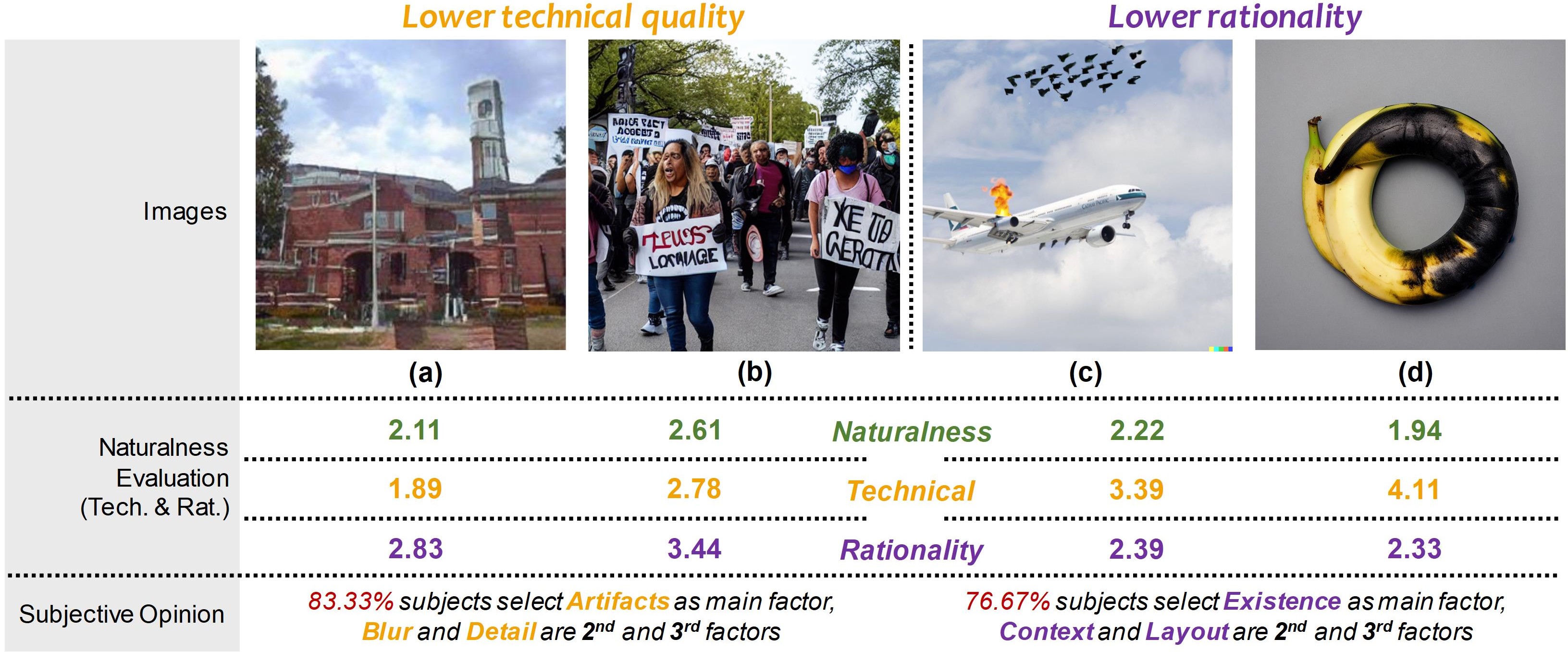}
   \caption{{\bf The motivation of} naturalness assessment for AI-generated images: multi-perspective settings can effectively avoid the {\bf perceptual bias} on single absolute evaluation, and provide more accurate judgments to serve as downstream supervision.}
   \label{intro-example}
\end{figure}

\vspace{-0.7cm}

\section{Introduction}
\label{sec:intro}
\noindent
Recent advancements in deep generative models have sparked a new craze in Artificial Intelligence-Generated Images (AGIs), which have gained significant progress across various applications, including text-to-image generation \cite{nichol2021glide, saharia2022photorealistic, ramesh2022hierarchical, rombach2022high, midj, dreamlike}, image translation \cite{zhang2020cross,patashnik2021styleclip,zhan2022bi, kwon2022diffusion,jiang2023masked}, image inpainting \cite{lugmayr2022repaint, li2022mat}, image colorization \cite{kang2022ddcolor, xia2022disentangled, wang2023unsupervised}, and image editing \cite{pan2023drag, brooks2023instructpix2pix, zhang2023magicbrush}.
However, even cutting-edge models occasionally generate irrational content or technical artifacts in the image, which we refer to as the image naturalness problem. 
Unlike natural scene images (NSIs) that are captured from real-world scenes, AI-driven image generation harnesses neural networks to learn synthesis rules from extensive image datasets \cite{lu2023seeing,wu2023ai,cao2023comprehensive}. Its instability and randomness of generation mode attach AGIs with more diverse content, leading to varying degrees of naturalness, which often requires retouching and filtering before practical use so as to avoid misleading people and negative social repercussions. Consequently, objective models for evaluating the naturalness of AGIs are urgently needed.

Conventionally, image naturalness is described as the degree of correspondence between a real-life scene and a photograph displayed on a device based on some technical criteria (\eg, \textit{texture}, \textit{exposure}, \textit{color reproduction}, \textit{shooting artifacts}) \cite{de1995naturalness,cadfk2005naturalness,choi2009investigation}, which has been utilized for image quality assessment (IQA) to compare and guide the optimization of systems and algorithms \cite{gu2016blind,yan2019naturalness,le2020study}.
Under this theory, the images with richer details (Fig. \ref{intro-example}(c) and Fig. \ref{intro-example}(d)) should have notably better naturalness than the blurred image in Fig. \ref{intro-example}(a), which is opposite to the human opinion.
More recently, the emergence of AGI broadens the definition of image naturalness to comprise more non-technical semantic factors (\eg, \textit{existence}, \textit{layout}), which are normally regarded as rationality perspective \cite{lu2023seeing,li2023image}. 
However, it is highly subjective and its mechanism of how rationality affects human perception in image naturalness reasoning, \textit{i.e., human-naturalness opinions}, is still ambiguous and may be multidimensionally coupled.

In this paper, we make the first attempt to evaluate the naturalness of AI-generated images, a new field of quality assessment with increasing attention \cite{otani2023toward,li2023towards,wu2023ai,zhang2023perceptual,chen2023quantifying}. 
 To benchmark the naturalness of AI-generated images, we contribute \underline{A}I-\underline{G}enerated \underline{I}mage \underline{N}aturalness (\textbf{AGIN}) database, the {\it first-of-its-kind} database to study this problem. Specifically, AGIN contains 6,049 images collected from five different generative tasks with 18 model variants to ensure diversity. 
A total of 907,350 human opinions for technical and rationality perspectives as well as their effects on overall naturalness scores were collected from 30 participants. Since the perspectives and factors studied in our research are related to common IQA problems and not limited to AI-generated images, our methods and insights also can be applicable to other forms of multimedia.

AGIN provides several valuable observations for understanding human reasoning in visual naturalness. Firstly, we find both technical distortions (\eg, \textit{contrast}, \textit{blur}, and \textit{generative artifacts}) and rationality distortions (\eg, \textit{existence}, \textit{color}, and \textit{layout}) can affect visual naturalness significantly. 
The proportion of technical and rationality factors for AGIs with worse naturalness scores ($\mathrm{MOS}\in[1,3]$) is about 1:1.17.
Secondly, we notice that most factors in the two perspectives are relevant, but have disparate impacts on the naturalness score, which can result in a biased naturalness assessment. 
Furthermore, we also observe that the overall naturalness score can be well-approximated by a linear weighted sum of technical score and rationality score ($\mathrm{MOS}=0.145\mathrm{MOS_T}+0.769\mathrm{MOS_R}$). This correlation suggests joint learning of technical and rationality branches can be a feasible way to predict naturalness.

With the AGIN database, we propose the \underline{J}oint \underline{O}bjective \underline{I}mage \underline{N}aturalness evalua\underline{T}or (\textbf{JOINT}), an objective naturalness assessment method that offers high alignment with human perception. JOINT aims to mimic human reasoning of image naturalness by jointly learning on both technical and rationality branches. Specifically, given the different characteristics of each branch, we elaborate several designs, such as \textit{patch partition}, \textit{deep feature regularization}, and \textit{pretraining}, to allocate each branch with corresponding learning interests. Two different supervision schemes including using the overall naturalness scores (JOINT) and the respective scores for each perspective (JOINT++) are applied to train the model.
Finally, we use an effective subjective weighting strategy combined with the predictions of two branches to compute the overall naturalness score.
Experimental results on AGIN database verify the effectiveness of our proposed JOINT and JOINT++ that not only outperform baselines on the overall naturalness assessment but also provide more subjectively consistent results for technical and rationality perspectives.

Our {\bf contributions} can be summarized as follows:
\begin{itemize}
    \item We take the first step to explore the naturalness of AI-generated images, focusing on five prevalent generative tasks. Our methods and findings can be applied to other forms of AI-generated multimedia.
    \item We contribute AGIN database, the first database that facilitates studying the naturalness of AI-generated images via human ratings on overall naturalness scores as well as the technical and rationality perspectives.
    \item Based on AGIN, we elucidate the mechanisms underlying human perception of image naturalness, providing insights into how technical and rationality factors influence human reasoning.
    \item We propose the JOINT, an objective naturalness evaluator for AI-generated images that models human perception on naturalness by a brain-inspired joint learning from technical and rationality perspectives, resulting in better performance.
\end{itemize}

\section{Related Work} 
\noindent
{\bf AI-generated Image and Naturalness.} 
Generative models have emerged as an effective paradigm for image synthesis \cite{rombach2022high,saharia2022photorealistic,nichol2021glide,ramesh2022hierarchical,midj,zhang2023adding}.
Nonetheless, most generative adversarial network (GAN)-based methods \cite{zhang2020cross,patashnik2021styleclip,zhan2022bi} are prone to produce visually unnatural results due to their instability and mode collapse issues. Even state-of-the-art diffusion-based generative models \cite{kwon2022diffusion,xia2022disentangled,lugmayr2022repaint,brooks2023instructpix2pix} introduce oftentimes perceptible unnatural perturbations such as spurious details, disordered layout, and color mismatch on images.
Prior naturalness prediction approaches \cite{gu2016blind,liu2019unsupervised,yan2019naturalness,guo2021underwater,zheng2022uif}, driven by image statistic distribution, have predominantly focused on natural scene images (NSIs), which fail in AGIs, where exist diverse contextual content variations with less significant intrinsic properties (\eg {\it resolution}, {\it color space}, and {\it image format}). 
As a result, it is challenging to design an effective naturalness assessment method for AI-generated images that can be used to optimize the naturalness of the generated images and make them more robust to real-world applications.

\noindent
{\bf AI-generated Image Assessment.}
Existing AI-generated image assessment research mainly focuses on perceptual quality.
Early objective metrics such as Inception Score (IS)  \cite{salimans2016improved} measures perceptual quality by calculating the uniformity of AGI group features from the output of Inception model.
Distance-based methods such as Fréchet Inception Distance (FID) \cite{heusel2017gans} and Kernel Inception Distance (KID) \cite{binkowski2018demystifying} as well as Precision-Recall \cite{kynkaanniemi2019improved} evaluate the discrepancy between distributions of AGI and NSI.
Nevertheless, the above methods are all \textit{group-targeted} and not suitable for assessing single image.
Besides, the widely used CLIPScore \cite{hessel2021clipscore} is already saturated in comparing state-of-the-art generative models with authentic images and can inflate for a model trained to optimize text-to-image alignment in the CLIP space \cite{otani2023toward}.
This empirical evidence of the failure of the automatic measures motivates human evaluation of perceived quality.
Kirstain \etal \cite{kirstain2023pick} collected human preferences between two generated images from the same prompt for text-to-image tasks.
Wang \etal \cite{wang2023aigciqa2023} investigated the impact of quality, fidelity, and correspondence of AI-generated images on human visual perception. Similarly, Li \etal \cite{li2023agiqa} conducted a subjective evaluation to annotate images from both perception and alignment dimensions with varying input prompts and internal parameters in AGI models.
However, these studies, which merely collect coarse, single-voice, and overall subjective opinions, lack the exploration of specific factors with fine-grained and explainable evaluations on various generation tasks. 

%------------------------------------------------------------------------
\begin{table*}
\setlength{\belowcaptionskip}{-0.01cm}
  \centering
\caption{Comparisons between the {\bf AGIN} database and existing IQA databases.}
  \renewcommand\arraystretch{1}
   \resizebox{\linewidth}{!}{\begin{tabular}{lccccc}
    \toprule[1pt]
     Database & Image Source& \#Content& \#Image & Perspective & Distortion \\
    \midrule
     LIVE (2004) \cite{sheikh2005live} & Kodak test set& 30& 779& Quality&  5 Artificial \\
     TID2008 (2008)\cite{ponomarenko2009tid2008}& Kodak test set & 25& 1,700&  Quality& 17 Artificial \\
     CSIQ (2009)\cite{larson2010most}& Kodak test set & 30& 866&  Quality& 6 Authentic \\
     TID2013 (2013)\cite{ponomarenko2015image} & Kodak test set& 25& 3,000& Quality& 24 Artificial \\
     LIVEC (2015)\cite{ghadiyaram2015massive} & Camera& 1162& 1162&  Quality& 15 Authentic \\
     WED (2017)\cite{ma2016waterloo} & Internet& 4,744& 94,880& Quality& 4 Artificial \\
     MDID (2017)\cite{sun2017mdid} & Internet& 20& 1,600&  Quality& 5 Artificial \\
     PieAPP (2018)\cite{prashnani2018pieapp} &WED \cite{ma2016waterloo}& 200& 20,280& Quality& 75 Artificial\\
     KADID-10k (2019)\cite{lin2019kadid}& Internet& 81& 10,125&  Quality& 25 Artificial\\
     KonIQ-10k (2020)\cite{hosu2020koniq} & Multimedia& 10,073& 10,073& Quality& $-$ Authentic \\
     PIPAL (2020)\cite{jinjin2020pipal} &DIV2K \cite{agustsson2017ntire}, Flickr2K \cite{timofte2017ntire}&250&29,000&Quality&40 Artificial\\
     PAN (2023)\cite{li2023towards} & Autonomous Driving& 2,688& 2,688& Naturalness& $-$ Adversarial \\
     AGIQA-1k (2023) \cite{zhang2023perceptual}&AI-generated&1,080&1,080&Quality&2 Generative\\
     AIGCIQA2023 (2023) \cite{wang2023aigciqa2023} &AI-generated&2,400&2,400&Quality, Authenticity, Correspondence&6 Generative\\
    AGIQA-3k (2023) \cite{li2023agiqa}&AI-generated&2,982&2,982&Quality, Alignment&6 Generative\\
    \midrule
     \bf AGIN (Ours) & \bf AI-generated& \bf 6,049& \bf 6,049& \bf Technical, Rationality, Naturalness&\bf 18 Generative\\
    \bottomrule[1pt]
  \end{tabular}}
  \label{database}
  \vspace{-2.0em}
\end{table*}

\section{AI-Generated Image Naturalness Database}
\label{sec::database}

\begin{figure}[t]
\setlength{\belowcaptionskip}{0.1cm}
\setlength{\belowcaptionskip}{-0.6cm}
  \centering
   \includegraphics[width=1\linewidth]{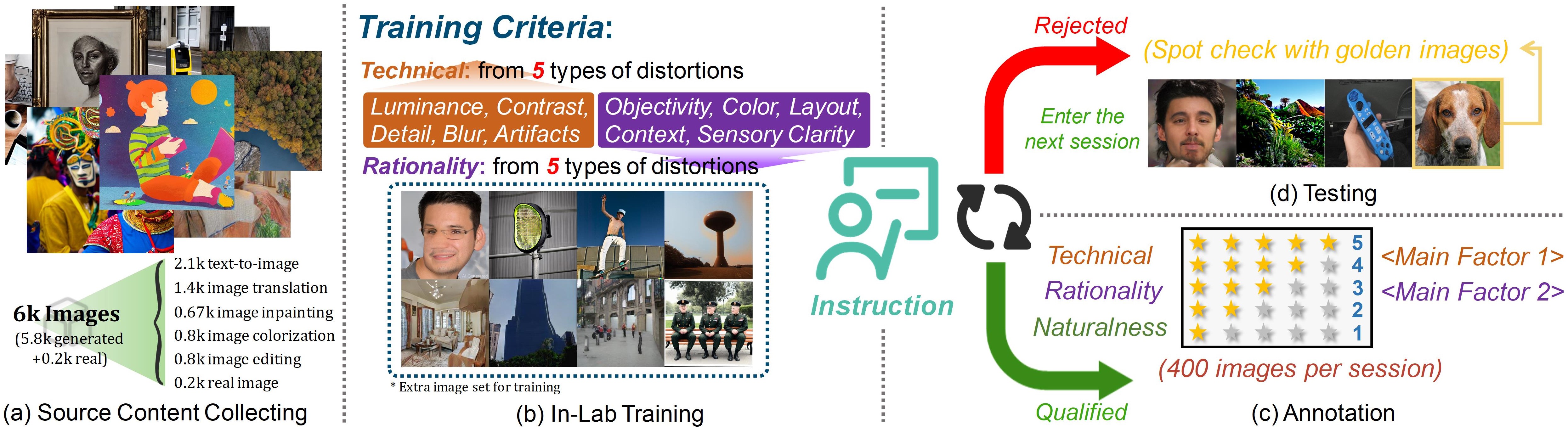}
   \caption{{\bf Workflow of the human evaluation in AGIN:} source images are first collected from 5 generative tasks and real-world image datasets {\bf (a)}, and then we conduct in-lab training with instructions {\bf (b)}. After that, subjects are asked to rate the images from three aspects {\bf (c)}, while carrying out tests {\bf (d)} to control the annotation quality.}
   \label{workflow_sub}
\end{figure}

\noindent
In this section, we elaborate on the construction procedures of the proposed AGIN database, along with the subjective human evaluation (Fig. \ref{workflow_sub}).
The database includes 6,049 AI-generated images, upon which we collected 907,350 ratings in terms of the overall naturalness score and its two perspectives: the technical and rationality scores, as well as their respective main influencing factor. A quick comparison of related datasets can be found in Tab. \ref{database}.

\subsection{Data Preparation}
\noindent
%{\bf Evaluated Baselines.}
As an initial investigation, we choose five sources of AI-generated images from text-to-image, image translation, image inpainting, image colorization, and image editing tasks, which typically suffer from naturalness problems.
We select 18 models including: (1) \textit{text-to-image}, 5 models with over 400 prompts are used for image generation. (2) \textit{image translation}, 5 models with various text-, image-, or mask-guided (\eg edge map, semantic map) translation. (3) \textit{image inpainting}, 2 models that take mutilated images as inputs. (4) \textit{image colorization}, 3 models that colorize the grayscale images. (5) \textit{image editing}, 3 models that perform layout or content editing on the image via text prompts or interactive anchor points. In addition to the AI-generated images, we also added 200 extra real images into the AGIN database to help analyze the accuracy of the subjective experiment and objective algorithms, which stand for a high level of naturalness. The distribution of the selected models and categories is displayed in the pie chart of Fig. \ref{intro}. \textit{We carefully reproduce the generation processes in supplementary materials}.

%--------------------------------------------------------------------------

%--------------------------------------------------------------------------

\subsection{Design of the Human Evaluation}

\subsubsection{Choice of Naturalness-related Factors.} 
We conduct the naturalness evaluation from two perspectives, \textit{i.e.}, the low-level technical perspective and the high-level rationality perspective, as follows.

\noindent
{\it Factors in Technical Perspective.} We consider specific image attributes (\eg luminance and contrast) that have high correlations of naturalness \cite{cadfk2005naturalness,gu2016blind,yan2019naturalness}. Besides, \textit{in-capture} authentic distortions \cite{yu2019predicting,beghdadi2022benchmarking}, such as reproduction of details and blur that happen in nature scene images, are considered.
Concretely, we study four distortions: 

\noindent
(T-1) \textit{Luminance}: Unrecognizable regions due to extremely high/low brightness.

\noindent
(T-2) \textit{Contrast}: High contrast produces a clearer and more vivid image, whereas low contrast leads to less color variety.

\noindent
(T-3) \textit{Detail}: Whether the image has detail or texture, such as wrinkles in clothing, hair, or skin.

\noindent
(T-4) \textit{Blur}: Clearness of image. Whether it is blurry or clear.

\noindent
and a common error introduced by the instability and mode collapse issues of generative models:

\noindent
(T-5) \textit{Artifacts}: Content discontinuity or meaningless objects \cite{zhang2023perceptual,zhu2023genimage,lu2023seeing,li2023agiqa}.

\noindent
{\it Factors in Rationality Perspective.} Compared to real images, AGIs possess richer content with diverse styles. Beyond technical distortions, the visual naturalness of AGIs is largely affected by rationality-related factors \cite{zhu2023genimage,lu2023seeing,chen2023x}. Such high-level factors are vaguely described as \textit{the memory of the real-life scene} in previous research \cite{de1995naturalness,de1996naturalness,choi2009investigation,le2020study}, which are not suitable for qualitative and quantitative analysis. 
In this work, we contribute 5 rationality dimensions to facilitate subjects to better rate their feeling on images.

\noindent
(R-1) \textit{Existence}: Whether the scene or objects in the image exist or could exist in the real world.

\noindent
(R-2) \textit{Color}: Does the image follow the natural color rule and present harmonious and pleasing colors?

\noindent
(R-3) \textit{Layout}: Is the image layout logical?

\noindent
(R-4) \textit{Context}: Whether the objects in the image are related.

\noindent
(R-5) \textit{Sensory Clarity}: The abstract perception. Whether the image content is easy to understand.

%\bf{Design of the Study.}
%\label{design_Sub}

\noindent
{\bf Participants and Apparatus.}
To ensure the comprehensiveness and reliability of the evaluation, we recruited 30 participants (18 male, 12 female, age=22.6±3.1) from campus, all with normal (corrected) eyesight. We conduct the subjective studies in-lab to ensure that all subjects have a clear and consistent understanding of all factors. Each participant is compensated \$240 for evaluating 6,049 images. All images are displayed on a 27-inch screen with a resolution of 2560$\times$1440 and a viewing distance of about 70cm. Note that we have addressed the ethical challenges involved in constructing such a database, by obtaining from each subject depicted in the database a signed and informed agreement, making it equipped with such legal and ethical characteristics.

\noindent
{\bf Rating Strategy and Wording.}
We discuss the concrete form for human evaluation as follows. 1) \textit{Task-oriented absolute choice}. Since the wording of questions and labels can significantly affect annotators’ labeling behavior, we abandon the traditional 3-point or 5-point Likert scale that only provides endpoint labels from \textit{worst} to \textit{best} and is too vague to describe the degree of naturalness. As a solution, we elaborate the questions and labels for three different perspectives to reduce subjectivity rather than using general ones (\eg \textit{bad}, \textit{good}, or \textit{very good}).
2) \textit{Pick up the main factor}. Most existing subjective studies merely focus on the assessment of the overall score but neglect to explore the underlying factors. Therefore, we ask subjects to choose a primary factor that affects most for each perspective after rating the general scores, which enables us to investigate the correlation between each dimension and image naturalness.

\noindent
{\bf Training, Testing, and Annotation.} The workflow of the human evaluation is illustrated in Fig. \ref{workflow_sub}. Before conducting the formal study, we manually generated 10 exemplars beyond the AGIN database for each dimension as training images to familiarize subjects with the goal of this evaluation. Subsequently, we instruct the subjects to rate the technical quality, rationality, and overall naturalness of each image from $\left\{ 1,2,3,4,5\right\}$, and select the main factors that affect the technical quality and rationality most. For testing, we randomly insert 10 \textit{golden images} into each session as an inspection to ensure the quality of annotation. 
\textit{We defer more details of human evaluation and quality control to supplementary materials.}

%--------------------------------------------------------------------------
\begin{figure}[t]
\setlength{\abovecaptionskip}{0.1cm}
\setlength{\belowcaptionskip}{-0.5cm}
  \centering
   \includegraphics[width=1\linewidth]{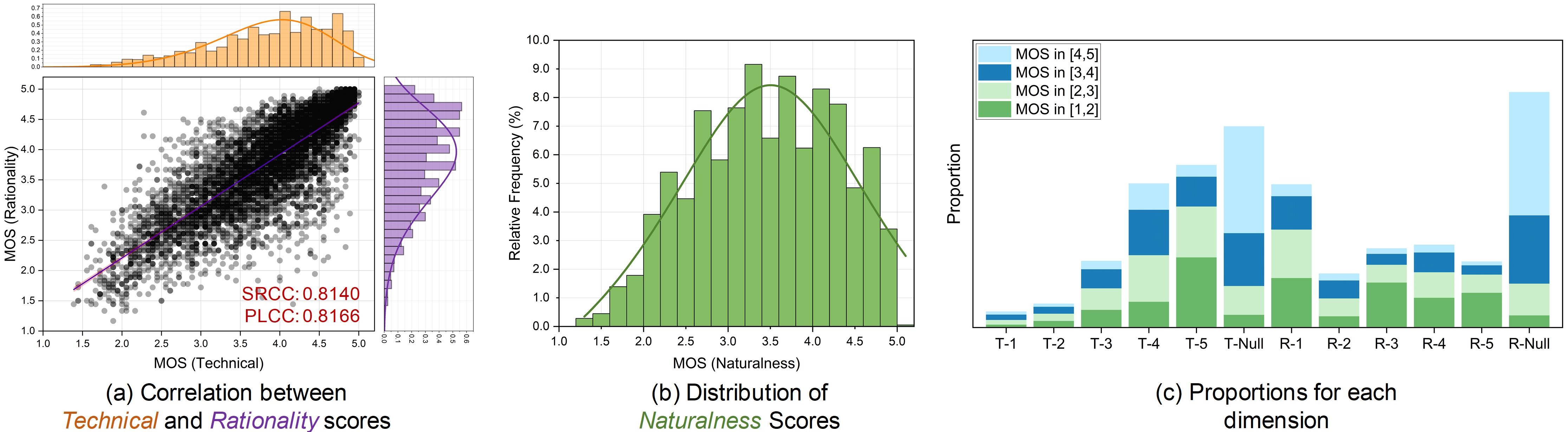}
   \caption{{\bf Data properties of AGIN.} (a) The correlations between technical and rationality perspectives, (b) distributions of overall naturalness scores, and (c) the tendency of main factors chosen by participants across different ranges of naturalness scores.}
   \label{factor_analysis}
\end{figure}
%--------------------------------------------------------------------------
%--------------------------------------------------------------------------
\begin{figure}[t]
\setlength{\abovecaptionskip}{0.1cm}
\setlength{\belowcaptionskip}{-0.65cm}
  \centering
   \includegraphics[width=1\linewidth]{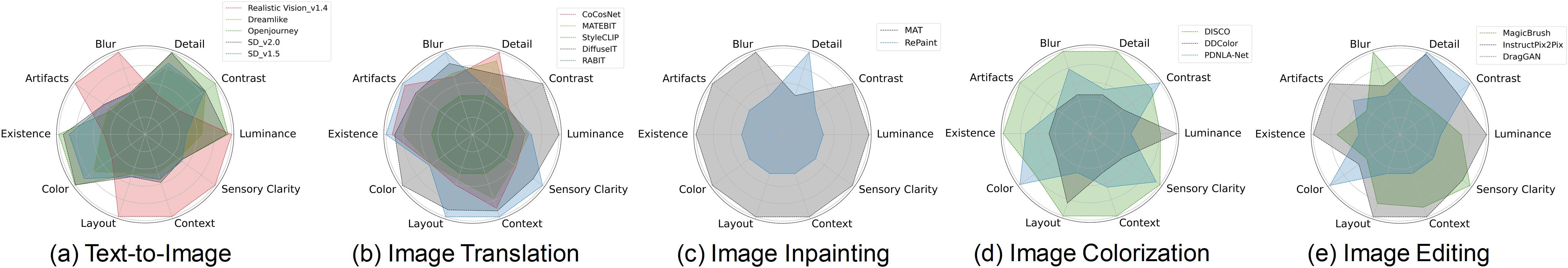}
   \caption{More comparisons of image generation models in terms of naturalness-related factors. Zoom-in for better visualization.}
   \label{radar_factor}
\end{figure}
%--------------------------------------------------------------------------

\subsection{Insights}
\label{insight}
\noindent
What affects the naturalness of AGIs? What are the latent correlations among different factors? Based on AGIN, we provide the following two insights: 

\noindent
{\bf Insight \ding{182}}: \textit{Naturalness is affected by both low-level technical distortions and high-level rationality distortions.}

%-------------------------------------------------------------------------
\begin{figure*}[t]
\setlength{\abovecaptionskip}{0.1cm}
\setlength{\belowcaptionskip}{-0.3cm}
  \centering
   \includegraphics[width=1\linewidth]{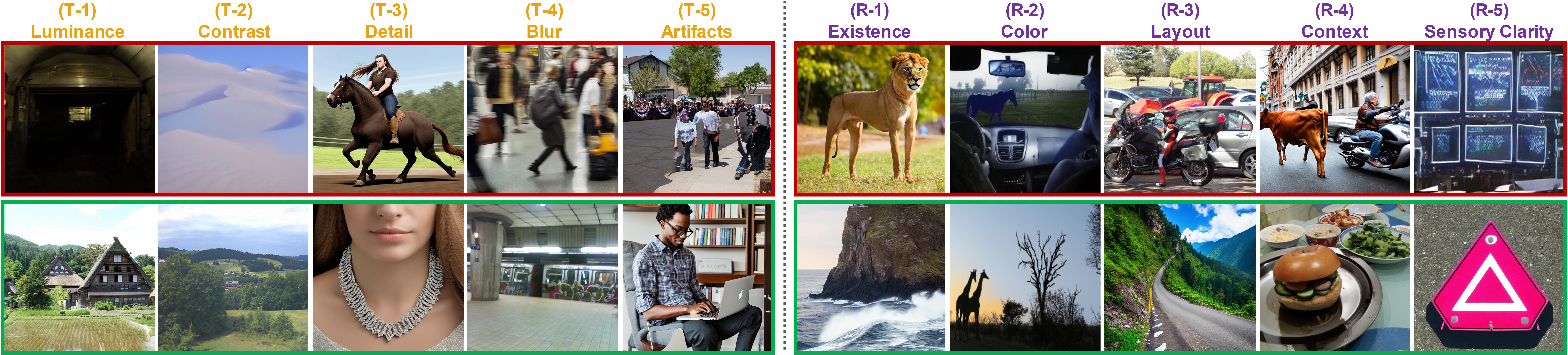}
   \caption{{Visualization of images with severe (1st row with red box) and minor effects (2nd row with green box) of each dimension.}}
   \label{factor_visual}
\end{figure*}

\begin{table}[t]
\setlength{\belowcaptionskip}{-0.01cm}
\caption{Correlation between different perspectives and overall naturalness in \textbf{AGIN}.}
  \centering
  \renewcommand\arraystretch{0.65}
  \begin{tabular}{p{1cm}<{\centering}p{1cm}<{\centering}p{1cm}<{\centering}cc}
    \toprule[1pt]
    \scriptsize Metrics &\scriptsize $\mathrm{MOS_T}$ &\scriptsize $\mathrm{MOS_R}$ &\scriptsize $\mathrm{MOS_T+MOS_R}$&\scriptsize $\mathrm{0.145MOS_T+0.769MOS_R}$ \\
    \midrule
    \scriptsize SRCC$\uparrow$ &\scriptsize 0.8647&\scriptsize 0.9694 &\scriptsize 0.9672&\scriptsize {\bf \color{red}{0.9777}} \\
    \scriptsize PLCC$\uparrow$ &\scriptsize 0.8599&\scriptsize 0.9639&\scriptsize 0.9580 &\scriptsize {\bf \color{red}{0.9713}}\\
    \bottomrule[1pt]
  \end{tabular}
  \label{correlation}
  \vspace{-2.0em}
\end{table}
%-------------------------------------------------------------------------

We first provide a visualization of the data properties in AGIN database (Fig. \ref{factor_analysis}\textcolor{red}{a}), from which we can observe the inner correlation between the technical and rationality perspectives.
Tab. \ref{correlation} lists the quantitative results of Spearman and Pearson correlation between different perspectives, where the mean technical score, mean rationality score, and mean naturalness score are denoted as $\mathrm{MOS_T}$, $\mathrm{MOS_R}$, and $\mathrm{MOS}$, respectively. It can be observed that the two perspectives affect naturalness unequally
(\ie, rationality has a greater impact on the overall naturalness than technical perspective, while the linear weighted sum of two perspectives can better approximate the overall naturalness than any single form). This could lead to biased naturalness evaluation unwittingly when using mainstream IQA models that follow the overall MOS regression strategy.

\noindent
{\bf Insight \ding{183}}: \textit{Factors in two perspectives are related, but have disparate impacts on the overall naturalness.}

Fig. \ref{factor_analysis}\textcolor{red}{c} shows the proportion of each factor in different ranges of naturalness scores, where \textit{T-Null} and \textit{R-Null} indicate that there exist no factors affecting naturalness or the subjects have difficulty in choosing the main factors. First, \textit{T-Null} and \textit{R-Null} are more prevalent in images with preferable naturalness ($\mathrm{MOS}\in[4,5]$), indicating the impact of technical and rationality perspectives on naturalness.
Additionally, we notice that humans are more sensitive to generated artifacts (T-5) and blur (T-4) in terms of technical quality while focusing more on the existence (R-1) of the image contents in terms of rationality. 
Furthermore, we find especial high proportions of artifacts (T-5), existence (R-1), and layout (R-3) in the case of poor naturalness ($\mathrm{MOS}\in[1,2]$), suggesting that they are important naturalness factors for AGIs to take into account. Considering the interrelation between technical quality and rationality, we speculate that severe artifact distortions may lead to irrational content and chaotic layout. 
For each generative task in AGIN, we calculate the occurrence frequency of each factor, as shown in Fig. \ref{radar_factor}.
We also provide examples with varying degrees of effect for each dimension in Fig. \ref{factor_visual}, to better illustrate the manifestations of the naturalness problem in different dimensions.
Overall, these newly contributed dimensions describe the naturalness concerns of AGIs, some of which have never been encountered in the conventional IQA domain, providing reliable intuitions for developing objective naturalness assessment models.

%-------------------------------------------------------------------------

%-------------------------------------------------------------------------

\begin{figure*}[t]
\setlength{\abovecaptionskip}{0.1cm}
\setlength{\belowcaptionskip}{-0.6cm}
  \centering
   \includegraphics[width=1\linewidth]{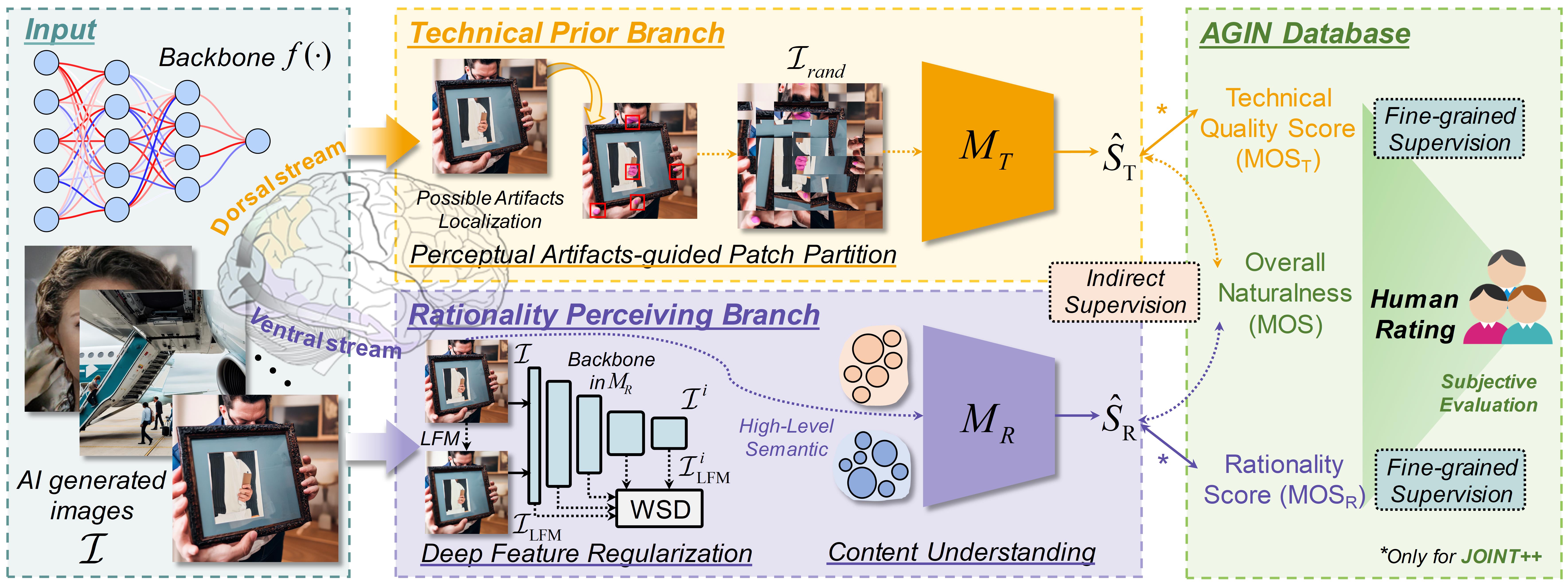}
   \caption{{\bf Framework of the proposed JOINT and JOINT++.} It consists of the technical prior branch (Sec. \ref{Branch_T}) and the rationality perceiving branch (Sec. \ref{Branch_R}) with indirect and fine-grained supervisions strategies (Sec. \ref{loss}).}
   \label{framework}
\end{figure*}

\vspace{-0.3cm}
\section{The Proposed JOINT and JOINT++}
\noindent
Studies in neurosciences \cite{ingle1973two,goodale1992separate,norman2002two} suggest that humans possess two distinct visual systems, which follow two main pathways, \textit{i.e.}, the dorsal stream and ventral stream, to handle low-level and high-level visual perception, respectively. To align model behavior with human perception process, we propose the \underline{J}oint \underline{O}bjective \underline{I}mage \underline{N}aturalness evalua\underline{T}or (\textbf{JOINT}) that models human naturalness reasoning by simultaneously considering the impact of technical and rationality perspectives with two independent branches, shown in Fig. \ref{framework}. To allocate these two branches with corresponding learning interests, we present several specific designs (\eg, \textit{patch partition}, \textit{feature regularization}) and two different training schemes, illustrated as follows.

%Specifically, the technical perspective mainly accounts for low-level distortions, \eg, \textit{texture losses}, \textit{blurs}, and \textit{artifacts} \cite{wu2023towards,wu2023exploring,sun2023blind,agarla2023quality}, while the rationality of image is largely related to high-level visual information including \textit{semantic}, \textit{attribute theme}, and \textit{layout} that are also the key elements of aesthetic assessment \cite{he2022rethinking,he2023thinking,li2020personality,yi2023towards}. Notably, some of the perceptual factors are related to both perspectives, such as \textit{contrast} that belongs to technical distortion and can affect \textit{color} harmonization, or \textit{artifacts} (a generative technical distortion but highly coupled with the \textit{existence} and \textit{layout} of the image contents). To allocate these two branches with corresponding learning interests, we present several specific designs (\eg, \textit{patch partition}, \textit{feature regularization}) and two different training schemes, illustrated as follows.
\vspace{-0.3cm}
\subsection{The Technical Prior Branch}
\label{Branch_T}
\noindent
For technical prior branch, we explicitly guide the model to prioritize the technical distortions while minimizing the impact of semantic information. Concretely, we randomly crop the image $\mathcal{I}$ into size-fixed patches and stitch them together ($\mathcal{I}_{rand}$) to disorganize most contents and layout while retaining technical distortions, thus destroying semantic information and rationality factors in images \cite{wu2022fast,wu2023exploring}. However, different from most global technical distortion, generated artifacts could become unrecognizable by random patch partition. Therefore, we propose to localize possible perceptual artifacts first and bypass these regions to keep their local distortion information. 

\noindent
{\bf Perceptual Artifacts-guided Patch Partition.}
Numerous research efforts have sought to localize the edited regions \cite{zheng2019survey,wang2019detecting}, which basically involves training a model to pinpoint systematic inconsistencies in generated images. More recently, Zhang \etal \cite{zhang2022perceptual,zhang2023perceptual} further expand this task to a fine-grained level that not only predicts inpainted areas but also detects and segments artifact areas that are noticeable to human perception. In this work, we use the detection model proposed in \cite{zhang2023perceptual} as the artifacts extractor to guide the patch partition. \textit{Note that we do not focus on the detection process in this work}. Given an image $\mathcal{I}$ of size $\mathcal{I}_W\times \mathcal{I}_H$, the patch partition can be formulated as:
\begin{align}
     m,n &= AExtractor(\mathcal{I}),\\
      {\mathcal{I}_{rand}} &= RPart\left( {{\mathcal{I}_{j \in [1,\frac{{{\mathcal{ I}_H}}}{N}]\backslash \{ m\},k \in [1,\frac{{{\mathcal{I}_W}}}{N}]\backslash \{ n\} }},{N_8}} \right),
\end{align}
where $AExtractor(\cdot)$ denotes the perceptual artifacts location extractor that returns coordinates for artifacts in $m$-th horizontal grid and $n$-th vertical grid. The divided patch size is $N\times N$. 
$RPart(\cdot, N_8)$ denotes a random partition within the 8-connected neighborhood of the patch, which destructs the local semantic information of the image while preserving the global semantics.

\vspace{-0.3cm}

\subsection{The Rationality Perceiving Branch}
\label{Branch_R}
\noindent
Since the high-level semantic information in rationality concerns is likewise of interest to the image aesthetic assessment (IAA), we pre-train this branch first with an IAA database and introduce a deep feature regularization to mitigate the effect of technical quality.

\noindent
{\bf Deep Feature Regularization.}
To maintain the principal content of the image and filter out the impact of partial technical factors, we use the piece-wise smooth algorithm \cite{Bar2006SemiblindIR} to obtain the low-frequency map of images $\mathcal{I}_{\mathrm{LFM}}$. 
Moreover, existing research \cite{zhang2018unreasonable, Liao2022DeepWSDPD} suggest that the distribution differences of deep features among different stages are related to technical distortions. Henceforth, we employ the one-dimension form of Wasserstein Distance (WSD) \cite{cazelles2020wasserstein} as a penalty constraint $\mathcal{L}_{\mathrm{WSD}}$ to eliminate the technical interference in rationality measuring by reducing the feature distribution divergence between $\mathcal{I}$ and $\mathcal{I}_{\mathrm{LFM}}$:
\begin{equation}
\setlength{\belowdisplayskip}{5pt}
{\mathcal{L}_{\mathrm{WSD}}} = {W_l}\left( {\mathcal{I},{\mathcal{I}_{\mathrm{LFM}}}} \right) + \sum\limits_{i = 1}^N {{W_l}\left( {{\mathcal{I}^i},\mathcal{I}_{\mathrm{LFM}}^i} \right)},
\end{equation}
where $\mathcal{I}^i$ and $\mathcal{I}^{i}_{\mathrm{LFM}}$ denote the extracted features of $\mathcal{I}$ and $\mathcal{I}_{\mathrm{LFM}}$ at the $i$-th stage. $W_l(\cdot,\cdot)$ is the Wasserstein distance with $l$-norm. 

%-------------------------------------------------

\subsection{Learning Objectives}
\label{loss}
\noindent
%We have discussed in Sec. \ref{insight} that the overall naturalness score $\mathrm{MOS}$ can be approximated as a weighted sum of $\mathrm{MOS_T}$ and $\mathrm{MOS_R}$. 
We propose to optimize two branches using the overall naturalness $\mathrm{MOS}$ via indirect supervision ($\mathcal{L}_{\mathrm{IS}}$).
However, the subjective bias between two perspectives can cause large absolute prediction errors and reduce the prediction accuracy for each branch. Hence, we add the Spearman Rank-order Correlation Coefficient (SRCC) loss \cite{li2020norm, Li2022BlindlyAQ,wu2023exploring} as a restraint to boost the prediction monotonicity of models. Overall, JOINT learns to assess image naturalness by minimizing:
\begin{align}
    \setlength{\abovedisplayskip}{0pt}%调整公式上方与正文间距
	\setlength{\belowdisplayskip}{0pt}%调整公式下方与正文间距
    {\mathcal{L}_{{\mathrm{IS}}}} = {\mathcal{L}_{{\mathrm{C}}}}\left( {{{\hat S}_{\mathrm{T}}},{\mathrm{MOS}}} \right) &+ {\mathcal{L}_{{\mathrm{C}}}}\left( {{{\hat S}_\mathrm{R}},{\mathrm{MOS}}} \right) + \beta {\mathcal{L}_{{\mathrm{WSD}}}}, \\
    {\mathcal{L}_{{\mathrm{C}}}} = {\mathcal{L}_{{\mathrm{MSE}}}} &+ \alpha {\mathcal{L}_{{\mathrm{SRCC}}}},
\end{align}
where $\alpha$ and $\beta$ are hyperparameters to control the strength of $\mathcal{L}_{\mathrm{SRCC}}$ and $\mathcal{L}_{\mathrm{WSD}}$, respectively. ${\hat S}_\mathrm{T}$ and ${\hat S}_\mathrm{R}$ denote the predicted score of technical and rationality branch.
Besides, based on the AGIN database, we also propose a fine-grained version ($\mathcal{L}_{\mathrm{FS}}$) using the corresponding perspective opinions for both branches:
\begin{equation}
    {\mathcal{L}_{{\mathrm{FS}}}} = {\mathcal{L}_{\mathrm{C}}}\left( {{\hat S}_\mathrm{T}},{\mathrm{MOS_T}} \right) + {\mathcal{L}_{\mathrm{C}}}\left( {{{\hat S}_\mathrm{R}},{\mathrm{MOS_R}}} \right),
\end{equation}
and the proposed JOINT++ is trained by a combination of the above two losses so as to obtain more accurate predictions for both branches:
\begin{equation}
{\mathcal{L}_{\mathrm{JOINT++}}} = {\mathcal{L}_{{\text{FS}}}} + \lambda_{\mathrm{IS}} {\mathcal{L}_{{\text{IS}}}}
\end{equation}

%-------------------------------------------------
\noindent
{\bf Subjective Weighting Strategy.} According to the subjective study in AGIN, we adopt a simple but effective weighting strategy to compute the overall naturalness prediction ($\hat{S}_\mathrm{N}$) from two perspectives: $\hat{S}_\mathrm{N}=0.145\hat{S}_\mathrm{T}+0.769\hat{S}_\mathrm{R}$. With better performance in experiments (Tab. \ref{ablation2}), this strategy further demonstrates the observations in Sec. \ref{insight}.  
%-------------------------------------------------------------------------

\section{Experiments}

\subsection{Experimental Settings}
\noindent
{\bf Databases and Baselines.}
We conduct experiments on our proposed AGIN database and select 15 state-of-the-art methods as references to be compared against, including two traditional NR-IQA methods: BRISQUE \cite{mittal2012no} and NIQE \cite{mittal2012making}; five deep NR-IQA methods: DBCNN \cite{zhang2018blind}, HyperIQA \cite{su2020blindly}, MUSIQ \cite{ke2021musiq}, UNIQUE \cite{zhang2021uncertainty}, and MANIQA \cite{yang2022maniqa}; four deep image aesthetic assessment (IAA) methods: PAIAA \cite{li2020personality}, TANet \cite{he2022rethinking}, Delegate Transformer \cite{he2023thinking}, and SAAN \cite{yi2023towards}; four contrastive language-image pre-training (CLIP) model-based IQA methods: CLIP-IQA \cite{wang2023exploring}, CLIP-IQA$^{+}$ \cite{wang2023exploring}, LIQE \cite{zhang2023blind}, and InternLM-XComposer \cite{zhang2023internlm}.

\noindent
{\bf Implementation Details.}
In the technical branch, we crop patch at size $64\times 64$, and Swin Transformer \cite{liu2021swin} is used as backbone. We use the ResNet50 backbone \cite{he2016deep} pre-trained with AVA \cite{murray2012ava} in the rationality branch. $\alpha$ is set as $1$. $\beta$ and $\lambda_{\mathrm{IS}}$ are set as $0.5$. We train our model for 30 epochs using the Adam optimizer \cite{kingma2014adam} with learning rate $2\times 10^{-5}$. The batch size is set to $32$. All experiments are conducted under 5 train-test splits.

%---------------------------------------
\begin{table*}[t]\footnotesize
\setlength{\belowcaptionskip}{-0.01cm}
\belowrulesep=0pt
\aboverulesep=0pt
  \caption{{\bf Validating the necessity of AGIN database.} All baselines are trained using datasets from their respective domains. \textbf{\textcolor{red}{Red}}, \textcolor{blue}{\textbf{Blue}}, and \underline{\textbf{Black}} indicate the best, second, and third best performance, respectively.}
  \centering
  \resizebox{\linewidth}{!}{\begin{tabular}{lcccccccc}
    \toprule[1pt]
    \multirow{2}{*}{\bf Methods}&\multirow{2}{*}{\bf Type}& \multirow{2}{*}{\bf \makecell{Pre-computed Statistics/\\ Pre-training Datasets}}&\multicolumn{2}{c}{\bf Technical}&\multicolumn{2}{c}{\bf Rationality}&\multicolumn{2}{c}{ \bf Naturalness}\\
    \cmidrule(lr){4-5} \cmidrule(lr){6-7} \cmidrule(lr){8-9} 
    &&& SRCC$\uparrow$& PLCC$\uparrow$& SRCC$\uparrow$& PLCC$\uparrow$& SRCC$\uparrow$& PLCC$\uparrow$\\
    \midrule
     BRISQUE (TIP, 2012) \cite{mittal2012no}&\multirow{2}{*}{\textit{\makecell{Traditional IQA\\ (handcraft features)}}}& KonIQ-10k \cite{hosu2020koniq}& 0.3544& 0.3602& 0.1268& 0.1299& 0.1618& 0.1660\\
     NIQE (SPL, 2013) \cite{mittal2012making}&& KonIQ-10k \cite{hosu2020koniq}& 0.1843& 0.1484& 0.0377& 0.0235& 0.0707& 0.0445 \\
    \hdashline 
     $^{\star}$DBCNN (TCSVT, 2018) \cite{zhang2018blind}&\multirow{11}{*}{\textit{\makecell{Deep IQA\\(deep features)}}}& TID2013 \cite{ponomarenko2015image}& 0.2664& 0.3138& 0.0888& 0.1199& 0.1209& 0.1376\\
    \multirow{2}{*}{ \textit{– – same as above – –}}&& LIVE Challenge \cite{ghadiyaram2015massive}& 0.4132 & 0.4903& 0.1518& 0.2082& 0.1993& 0.2422 \\
    &&KonIQ-10k \cite{hosu2020koniq}& 0.4951& 0.5252& 0.2275& 0.2492& 0.2786& 0.2956 \\
     HyperIQA (CVPR, 2020) \cite{su2020blindly}&& KonIQ-10k \cite{hosu2020koniq}& 0.4953& 0.5541& 0.2839& 0.3211& 0.3332& 0.3725 \\
     $^{\star}$ MUSIQ (ICCV, 2021) \cite{ke2021musiq}&& PaQ-2-PiQ \cite{ying2020patches}& 0.4329& 0.4709& 0.2061& 0.2399& 0.2443& 0.2799\\
    \multirow{2}{*}{ \textit{– – same as above – –}}&& KonIQ-10k \cite{hosu2020koniq} & 0.4817& 0.5262& 0.2512& 0.2847& 0.2951& 0.3271\\
    && SPAQ \cite{fang2020perceptual}& 0.4324& 0.5166& 0.2193& 0.2741& 0.2561& 0.3085\\
      UNIQUE (TIP, 2021) \cite{zhang2021uncertainty}&& ImageNet \cite{russakovsky2015imagenet}& 0.5178& 0.5756& 0.2912& 0.3324& 0.3339& 0.3735\\
     $^{\star}$MANIQA (CVPRW, 2022) \cite{yang2022maniqa}&& KADID-10k \cite{lin2019kadid}& 0.4154& 0.4214& 0.2733& 0.2655& 0.3003& 0.3001\\
   \multirow{2}{*}{ \textit{– – same as above – –}}&& KonIQ-10k \cite{hosu2020koniq}& {\underline{\textbf{0.5771}}}& {\underline{\textbf{0.5902}}}& {\underline{\textbf{0.3453}}}& {\underline{\textbf{0.3594}}}& {\underline{\textbf{0.3937}}}& {\underline{\textbf{0.4034}}}\\
    && PIPAL2022 \cite{gu2022ntire}& 0.4014& 0.4407& 0.1985& 0.2354& 0.2341& 0.2597\\
    \hdashline
     PAIAA (TIP, 2020) \cite{li2020personality}&\multirow{4}{*}{\textit{\makecell{Deep IAA\\ (deep features)}}}&PsychoFlickr \cite{cristani2013unveiling}& 0.1363& 0.1234& 0.2445& 0.2587& 0.2261& 0.2298 \\
     TANet (IJCAI, 2022) \cite{he2022rethinking} && TAD66k \cite{he2022rethinking}& 0.1894& 0.2015& 0.2774& 0.2803& 0.2530& 0.2619\\
     Delegate Transformer (ICCV, 2023) \cite{he2023thinking} && AVA \cite{murray2012ava}& 0.2549& 0.3142& 0.2348& 0.2278& 0.2232& 0.2583\\
     SAAN (CVPR, 2023) \cite{yi2023towards}& &BAID \cite{yi2023towards}& 0.0515& 0.0359& 0.1477& 0.1380& 0.1413& 0.1456 \\
    \hdashline
     CLIP-IQA (AAAI, 2023) \cite{wang2023exploring}&\multirow{4}{*}{\textit{\makecell{CLIP-based model\\ (visual language prior)}}}&WIT-400M& 0.2114& 0.3275& 0.0348& 0.0827& 0.0167& 0.1109 \\
     CLIP-IQA$^{+}$ (AAAI, 2023) \cite{wang2023exploring}& &WIT-400M+KonIQ-10k \cite{hosu2020koniq}& 0.4959& 0.5595& 0.2613& 0.3189& 0.3078& 0.3550\\
      LIQE (CVPR, 2023) \cite{zhang2023blind}& &\textit{hybrid} \cite{sheikh2005live,ciancio2010no,larson2010most,ghadiyaram2015massive,lin2019kadid,hosu2020koniq}& 0.4928& 0.5428& 0.2457& 0.2765& 0.2974& 0.3244\\
      InternLM-XComposer (arxiv, 2023) \cite{zhang2023internlm}&&Q-Instruct \cite{wu2023q} &0.4741&0.5085&0.3074&0.3271&0.3268&0.3566\\
    \hdashline
     {\bf JOINT} (Ours) &\multirow{2}{*}{\textit{\makecell{Deep INA\\ (deep features)}}}&AGIN\textsubscript{train}& \textbf{\textcolor{blue}{0.8173}}& \textbf{\textcolor{blue}{0.8235}}& \textbf{\textcolor{blue}{0.7564}}& \textbf{\textcolor{blue}{0.7711}}& \textbf{\textcolor{blue}{0.7986}}& \textbf{\textcolor{blue}{0.8028}}\\
     {\bf JOINT++} (Ours)& &AGIN\textsubscript{train}& \textbf{\textcolor{red}{0.8351}}& \textbf{\textcolor{red}{0.8429}}& \textbf{\textcolor{red}{0.8033}}& \textbf{\textcolor{red}{0.8127}}& \textbf{\textcolor{red}{0.8264}}& \textbf{\textcolor{red}{0.8362}}\\
    \bottomrule[1pt]
  \end{tabular}}
  \label{exp_nece}
  \vspace{-2em}
\end{table*}
%-----------------------------------------------------------

\subsection{Exploring the Necessity of AGIN Database}
\label{AGIN_nece}
\noindent
In this section, we conduct experiments to verify whether the existing IQA and IAA databases can solve the problem of AI-generated image naturalness assessment, \textit{i.e.}, the necessity of AGIN database. Specifically, we test the baselines on AGIN using their respective pre-trained models. 
We can obtain the following observations from Tab. \ref{exp_nece}: (1) Our AGIN database is of great importance for assessing the naturalness of AI-generated images. The proposed JOINT++ outperforms MANIQA \cite{yang2022maniqa}, the second-best method pre-trained on KonIQ-10k \cite{hosu2020koniq}, by $0.4327$ ($+109.91\%$) in SRCC and $0.4328$ ($+107.29\%$) in PLCC, which shows the inferiority of existing IQA, IAA, visual-language models as well as datasets in evaluating the naturalness of AI-generated images. 
(2) Evaluating images from technical and rationality perspectives exhibits significant differences. We notice that IQA methods show relatively better performance in evaluating technical quality, whereas IAA methods excel in assessing rationality, which underscores the necessity of our distinct exploration of each perspective in the AGIN database.  
(3) Image naturalness assessment is different from both quality assessment and aesthetic assessment. Since mainstream IQA and IAA approaches fail to provide subjectively consistent evaluation results for image naturalness (more than $105\%$ lower in SRCC/PLCC), we speculate that this is due to the diverse characteristics of the image sources and disparities in task objectives, illustrating the necessity of constructing AGIN and exploring the influencing factors. 

\vspace{-0.3cm}
\subsection{Evaluation on the AGIN}
\label{eval_AGIN}

\noindent
{\bf Quantitative Studies.}
We benchmark recent state-of-the-art IQA and IAA methods by conducting training and testing in the AGIN. As shown in Tab. \ref{AGIN_train_test}, the two classical IQA methods \cite{mittal2012no,mittal2012making} perform significantly worse than deep IQA methods, and the proposed JOINT++ still achieves the best performance in terms of technical, rationality, and overall naturalness assessment. Surprisingly, all IAA methods exhibit subpar performance that on average $44.74\%$/$45.56\%$ lower than JOINT++ in naturalness evaluation. Their ineffectiveness can be attributed to a lack of consideration for technical factors and an attention bias in understanding the semantics of the content itself. Besides, most IAA models aim to learn more about global information (\eg, \textit{semantic}, \textit{composition}) than local elements that could overwhelmingly affect naturalness. Furthermore, both IQA and IAA approaches solely consider a single perspective with highly coupled factors in image naturalness reasoning, thereby rendering them incapable of providing reliable results.

\begin{table}[t]\scriptsize
\setlength{\belowcaptionskip}{-0.01cm}
\belowrulesep=0pt
\aboverulesep=0pt
\renewcommand\arraystretch{1.1}
  \caption{{\bf Performance comparisons on the AGIN.} We retrained all models using the score of each corresponding perspective.}
  \centering
 \begin{tabular}{lp{1.2cm}<{\centering}p{1.2cm}<{\centering}p{1.2cm}<{\centering}p{1.2cm}<{\centering}p{1.2cm}<{\centering}p{1.2cm}<{\centering}p{1.2cm}<{\centering}}
    \toprule[1pt]
     \multirow{2}{*}{\bf Methods} &\multicolumn{2}{c}{\bf Technical}&\multicolumn{2}{c}{\bf Rationality}&\multicolumn{2}{c}{\bf Naturalness}\\
    \cmidrule(lr){2-3} \cmidrule(lr){4-5} \cmidrule(lr){6-7}
    & SRCC$\uparrow$& PLCC$\uparrow$& SRCC$\uparrow$& PLCC$\uparrow$& SRCC$\uparrow$& PLCC$\uparrow$\\
    \midrule
     BRISQUE \cite{mittal2012no}& 0.4867& 0.4909& 0.3608& 0.3684& 0.3745& 0.4067\\
     NIQE \cite{mittal2012making}& 0.4235& 0.4279& 0.3144& 0.3211& 0.3358& 0.3378 \\
    \hdashline  
     DBCNN \cite{zhang2018blind}& 0.7623& 0.7661& 0.6834& 0.6838& 0.7057& 0.7128\\
     HyperIQA \cite{su2020blindly}& 0.7752& 0.7806 & {\underline{\textbf{0.7196}}}& {\underline{\textbf{0.7292}}}& 0.7365& {\underline{\textbf{0.7509}}}\\
     MUSIQ \cite{ke2021musiq}& 0.7286& 0.7355& 0.6974& 0.7013& 0.7066& 0.7103\\
     UNIQUE \cite{zhang2021uncertainty}& 0.7358& 0.7434& 0.6583& 0.6685& 0.6772& 0.6789\\
     MANIQA \cite{yang2022maniqa} & {\underline{\textbf{0.7763}}}& {\underline{\textbf{0.7817}}}& 0.7192& 0.7217& {\underline{\textbf{0.7385}}}& 0.7343 \\
    \hdashline
     PAIAA \cite{li2020personality}& 0.4763& 0.4833& 0.4532& 0.4596& 0.4483& 0.4528\\
     TANet \cite{he2022rethinking}& 0.5367& 0.5587& 0.4731& 0.4762& 0.4782& 0.4535\\
     Del. Transf. \cite{he2023thinking} & 0.5882& 0.6134& 0.5037& 0.4942& 0.4805& 0.4961\\
     SAAN \cite{yi2023towards}& 0.4299& 0.4380& 0.4009& 0.4160& 0.4196& 0.4184\\
    \hdashline
     {\bf JOINT} (Ours) & \textbf{\textcolor{blue}{0.8173}}& \textbf{\textcolor{blue}{0.8235}}& \textbf{\textcolor{blue}{0.7564}}& \textbf{\textcolor{blue}{0.7711}}& \textbf{\textcolor{blue}{0.7986}}& \textbf{\textcolor{blue}{0.8028}} \\
     {\bf JOINT++} (Ours)& \textbf{\textcolor{red}{0.8351}}& \textbf{\textcolor{red}{0.8429}}& \textbf{\textcolor{red}{0.8033}}& \textbf{\textcolor{red}{0.8127}}& \textbf{\textcolor{red}{0.8264}}& \textbf{\textcolor{red}{0.8362}}\\
    \bottomrule[1pt]
  \end{tabular}
  \label{AGIN_train_test}
  \vspace{-1em}
\end{table}

%----------------------------------------------
\begin{figure}[t]
\setlength{\abovecaptionskip}{0.1cm}
  \centering
   \includegraphics[width=1\linewidth]{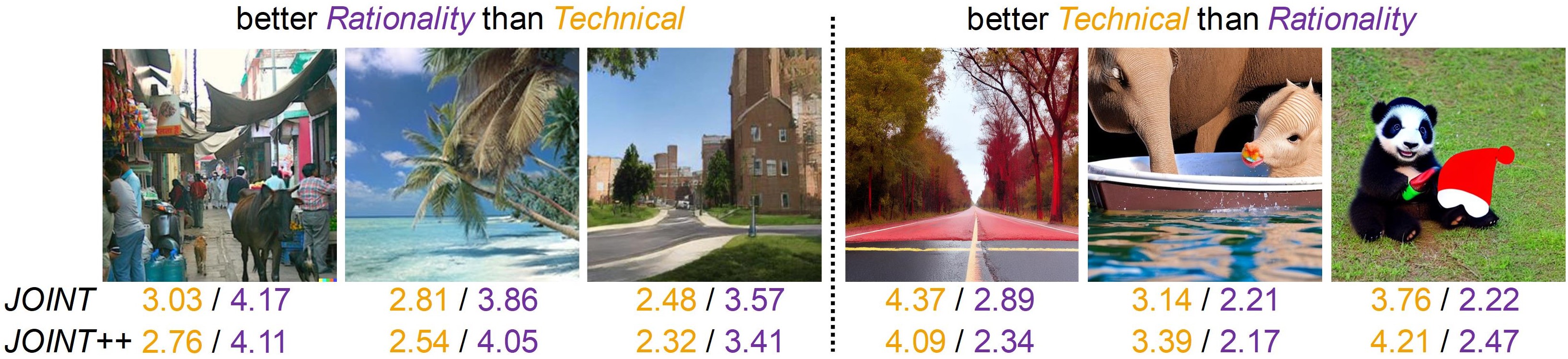}
   \caption{{\bf Qualitative Studies of JOINT and JOINT++.} Visualizations of images that have opposite technical quality and rationality. Zoom-in for better visualization.}
   \label{visual}
    \vspace{-2em}
\end{figure}
%----------------------------------------------

\noindent
{\bf Qualitative Studies.} As shown in Fig. \ref{visual}, we visualize two typical scenarios where the predicted technical and rationality scores are significantly diverged.
The images on the left with better rationality scores depict more realistic scenes yet suffer from \textit{noise}, \textit{blur}, \textit{artifacts}. In contrast, the images on the right with better technical scores are rich in details but with \textit{irrational colors}, \textit{nonexistent contents}, and \textit{irrelevant objects}. These observations align with human perception of the two perspectives, further substantiating the effectiveness and necessity of our joint learning strategy that can provide subjectively consistent image naturalness predictions.

%-------------------------------------------------

\begin{table}\scriptsize
\vspace{-2em}
\setlength{\belowcaptionskip}{-0.01cm}
\belowrulesep=0pt
\aboverulesep=0pt
\renewcommand\arraystretch{1.1}
  \centering
    \caption{{\bf Ablation study of JOINT (I):} the effects of specific designs.}
 \begin{tabular}{lp{1cm}<{\centering}p{1cm}<{\centering}p{1cm}<{\centering}p{1cm}<{\centering}p{1cm}<{\centering}p{1cm}<{\centering}}
    \toprule[1pt]
     \textbf{Perspective/} & \multicolumn{2}{c}{\textbf{Technical}}&\multicolumn{2}{c}{\textbf{Rationality}}&\multicolumn{2}{c}{\textbf{Naturalness}}\\
    \cmidrule{2-7}
     Variants/Metric& SRCC$\uparrow$&PLCC$\uparrow$&SRCC$\uparrow$&PLCC$\uparrow$&SRCC$\uparrow$&PLCC$\uparrow$\\
    \midrule
     \textit{w/o} Localization&0.811&0.816&0.755&0.769&0.782&0.794\\
     \textit{w/o} Regularization&0.814&0.820&0.729&0.738&0.758&0.766\\
     \textit{w/o} Multi-perspective&0.768&0.781&0.703&0.712&0.727&0.733\\
    \midrule
\textbf{JOINT} (Ours)& \textbf{\textcolor{red}{0.817}}&\textbf{\textcolor{red}{0.824}}& \textbf{\textcolor{red}{0.756}}&\textbf{\textcolor{red}{0.771}}& \textbf{\textcolor{red}{0.799}}&\textbf{\textcolor{red}{0.803}}\\
    \bottomrule[1pt]
  \end{tabular}
  \label{ablation1}
  \vspace{-4em}
\end{table}

%-------------------------------------------------

\begin{table}\scriptsize
\setlength{\belowcaptionskip}{-0.01cm}
\belowrulesep=0pt
\aboverulesep=0pt
\renewcommand\arraystretch{1.1}
  \centering
    \caption{{\bf Ablation study of JOINT (II):} correlation between perspectives and the effect of subjective weighting (denoted as $\oplus$).}
\begin{tabular}{p{0.8cm}<{\centering}p{0.8cm}<{\centering}p{0.8cm}<{\centering}p{1cm}<{\centering}p{1cm}<{\centering}p{1cm}<{\centering}p{1cm}<{\centering}p{1cm}<{\centering}p{1cm}<{\centering}}
    \toprule[1pt]
    \multicolumn{3}{c}{\textbf{Variants}} &\multicolumn{2}{c}{\textbf{Technical}}&\multicolumn{2}{c}{\textbf{Rationality}}&  \multicolumn{2}{c}{\textbf{Naturalness}}\\
    \midrule
     $\hat{S}_{\mathrm{T}}$& $\hat{S}_{\mathrm{R}}$&  $\oplus $&SRCC$\uparrow$&PLCC$\uparrow$&SRCC$\uparrow$&PLCC$\uparrow$&SRCC$\uparrow$&PLCC$\uparrow$\\
    \midrule
     $\checkmark $& & & \textbf{\textcolor{red}{0.817}}&\textbf{\textcolor{red}{0.824}}& 0.720&0.724& 0.725&0.744\\
     & $\checkmark $&& 0.687&0.699& \textbf{\textcolor{red}{0.756}}&\textbf{\textcolor{red}{0.771}}&0.767&0.763\\
    \midrule
     $\checkmark$&$\checkmark$&&0.753&0.768&0.732&0.744&0.759&0.755\\
 $\checkmark$&$\checkmark$&$\checkmark$&0.711&0.723&0.746&0.762& \textbf{\textcolor{red}{0.799}}&\textbf{\textcolor{red}{0.803}}\\
    \bottomrule[1pt]
  \end{tabular}
  \label{ablation2}
  \vspace{-2em}
\end{table}

%-----------------------------------------------------------
\subsection{Ablation Studies}
\label{ablation}
\noindent
{\bf Effects of Specific Designs.} In Tab. \ref{ablation1}, we verify the effect of three special designs in JOINT by keeping other parts the same.
First, it is superior to the variant \textit{w/o Localization} that randomly shuffles the patches and destructs the perceptual artifact regions, showing the necessity of preserving the local artifact distortion information. Secondly, with the deep feature regularization, JOINT is able to focus more on the rationality perspective.
Furthermore, JOINT is notably better than the variant \textit{w/o Multi-perspective} that directly takes the original images as inputs of both branches, proving the effectiveness of multi-perspective joint learning strategy.

\noindent
{\bf Effects of Subjective Weighting Strategy.} As reported in Tab. \ref{ablation2}, any single branch can not adequately represent the naturalness, and directly taking $\hat{S}_{\mathrm{T}}+\hat{S}_{\mathrm{R}}$ as overall naturalness
without weighting will also bring a notable performance decrease ($-5.01\%$/$-5.98\%$ in terms of SRCC/PLCC), which further supports the observations found in the AGIN.  

\noindent
{\bf Effects of Learning Objectives.} As shown in Tab. \ref{ablation3}, supervised by  corresponding MOS labels yields more accurate predictions. Compared with $\mathcal{L}_{\mathrm{IS}}$, $\mathcal{L}_{\mathrm{FS}}$ achieves around $+1.71\%$/$+1.21\%$, $+3.97\%$/$+2.98\%$, and $+2.50\%$/$+3.11\%$ performance gains in terms of technical, rationality, and naturalness assessment, respectively. 
It is also worth noting that combining $\mathcal{L}_{\mathrm{IS}}$ with $\mathcal{L}_{\mathrm{FS}}$ can significantly improve the prediction accuracy of overall naturalness.
Additionally, even without MOS labels for each perspective, $\mathcal{L}_{\mathrm{IS}}$ can still achieve comparable performance, which suggests the feasibility of modeling the human perception of naturalness from technical and rationality perspectives.

%-------------------------------------------------
\vspace{-0.5cm}
\begin{table}\scriptsize
\setlength{\belowcaptionskip}{-0.01cm}
\belowrulesep=0pt
\aboverulesep=0pt
\renewcommand\arraystretch{1.1}
  \centering
\caption{{\bf Ablation study of JOINT++:} the learning objectives.}
  %\setlength{\tabcolsep}{5pt}
  %\addtolength{\tabcolsep}{-1pt}
  \begin{tabular}{ccp{1cm}<{\centering}p{1cm}<{\centering}p{1cm}<{\centering}p{1cm}<{\centering}p{1cm}<{\centering}p{1cm}<{\centering}}
    \toprule[1pt]
    \multicolumn{2}{c}{\textbf{Loss Function}} &\multicolumn{2}{c}{\textbf{Technical}}&\multicolumn{2}{c}{\textbf{Rationality}}&\multicolumn{2}{c}{\textbf{Naturalness}}\\
    \midrule
     $\mathcal{L}_{\mathrm{IS}}$&$\mathcal{L}_{\mathrm{FS}}$&SRCC$\uparrow$&PLCC$\uparrow$&SRCC$\uparrow$&PLCC$\uparrow$&SRCC$\uparrow$&PLCC$\uparrow$\\
    \midrule
     $\checkmark $&&0.817&0.824&0.756&0.771&0.799&0.803\\
     & $\checkmark $&0.831&0.834&0.786&0.794&0.819&0.828\\
    \midrule
$\checkmark $&$\checkmark $& \textbf{\textcolor{red}{0.835}}&\textbf{\textcolor{red}{0.843}}& \textbf{\textcolor{red}{0.803}}&\textbf{\textcolor{red}{0.813}}& \textbf{\textcolor{red}{0.826}}&\textbf{\textcolor{red}{0.836}}\\
 \bottomrule[1pt]
  \end{tabular}
  \label{ablation3}
  \vspace{-4em}
\end{table}

%-------------------------------------------------

\section{Conclusion}
\noindent
In this paper, we contribute the AGIN database and the first subjective evaluation aimed at exploring the impact of technical and rationality perspectives on the naturalness of AGIs.
Besides, we propose JOINT, an objective naturalness evaluator that achieves higher alignment with human opinions against existing IQA and IAA approaches. 
Our work benefits the community by 1) presenting AGIN, which enables research on benchmarking and evaluating the naturalness of AGIs by multi-dimensional human ratings; 2) encouraging new research on the naturalness assessment of AGIs via analysis of technical and rationality features; 3) promoting the development of better INA algorithms for AGIs or other forms of AI-generated multimedia.

%\clearpage  % TODO REVIEW/FINAL: This \clearpage needs to be removed from both review and camera-ready versions.

\appendix

In this supplementary material, we provide additional details that we do not mention in the main paper.

\section{Extended Details of AGIN Database}
\begin{figure*}
  \centering
   \includegraphics[width=1\linewidth]{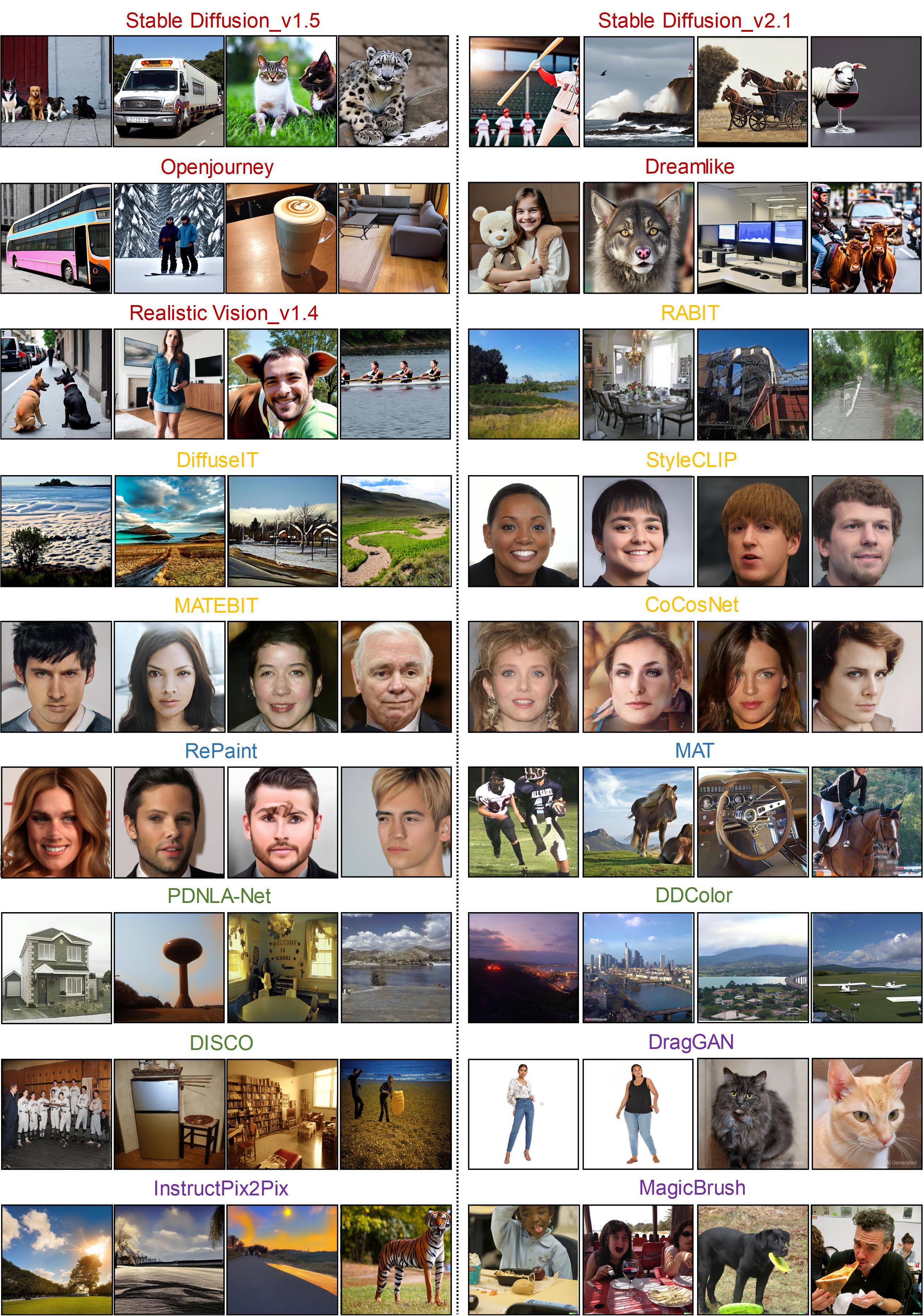}
   \caption{{\bf Additional results on generated images in AGIN database.} We show four examples for each model in five tasks (text-to-image, image translation, image inpainting, image colorization, and image editing).}
   \label{additionimg}
\end{figure*}

\noindent
In this section, we provide additional details for the AGIN database. 
We define the task of evaluating the naturalness of AI-generated images as a special case of image quality assessment (IQA). As shown in Tab. 1 from the main paper, AGIN differs from the existing IQA database in three aspects: image source, evaluation perspective, and distortion type. 
Previous IQA databases mainly focus on \textit{post-capture} artificial distortion (\eg, masked noise, JPEG compression) or \textit{in-capture} authentic distortions (\eg, motion blur, exposure), while AI-generated images possess entirely different generative patterns and richer contents, posing new challenges for constructing a comprehensive database.
The proposed AGIN provides guidance for redefining image quality in the AIGC field.
Note that since we focus on the naturalness assessment of AI-generated images, only prompts that describe objective objects or phenomena and source images with authentic style are used to build the database. It is unnecessary to add abstract or surreal concepts, which are prone to appear unnatural. To cover diverse appearances of naturalness issues, we collect multi-sourced images from five tasks, as depicted in Tab. \ref{appendix_model_diversity}. Exemplar images for each model are shown in Fig. \ref{additionimg}.

\begin{table*}
  \centering
  \renewcommand\arraystretch{0.9}
  \setlength{\belowcaptionskip}{-0.01cm}
  \caption{{\bf Detailed information of the models used in AGIN.} ``$\rhd$" indicates the upsampling operation.}
  \resizebox{\linewidth}{!}{\begin{tabular}{lccccc}
    \toprule[1.25pt]
     Baseline & Model& Generator& Resolution& \#Content& Input Source \\
    \midrule
     \multirow{5}{*}{Text-to-Image} & Stable Diffusion\_v1.5 (CVPR'22) \cite{rombach2022high}& Diffusion& $512^2$& 428& \multirow{5}{*}{\makecell{Prompts are from\\ COCO Caption \cite{lin2014microsoft}\\, DrawBench \cite{saharia2022photorealistic}\\, and ChatGPT.} } \\
     & Stable Diffusion\_v2.1 (CVPR'22) \cite{rombach2022high}& Diffusion& $512^2$& 428&    \\
      & Openjourney \cite{midj,rombach2022high}& Diffusion& $512^2$& 428&  \\
      & Dreamlike \cite{dreamlike,rombach2022high}& Diffusion& $768^2$& 428&  \\
      & Realistic Vision\_v1.4 \cite{rombach2022high}& Diffusion& $512^2$& 428&  \\
    \midrule
     \multirow{5}{*}{Image Translation}& RABIT (ECCV'22) \cite{zhan2022bi}& GAN& $256^2\rhd 512^2$& 500& ADE20K \cite{zhou2017scene} \\
      & DiffuseIT (ICLR'23) \cite{kwon2022diffusion}& Diffusion& $256^2\rhd 512^2$& 300& Landscape \cite{chen2018cartoongan} \\
      & StyleCLIP (ICCV'21) \cite{patashnik2021styleclip}& GAN& $1024^2$& 300& FFHQ \cite{karras2019style} \\
      & MATEBIT (CVPR'23) \cite{jiang2023masked}& GAN& $512^2$& 149& CelebA-HQ \cite{liu2015deep} \\
      & CoCosNet (CVPR'20) \cite{zhang2020cross}& GAN& $512^2$& 157& CelebA-HQ \cite{liu2015deep} \\
    \midrule
     \multirow{2}{*}{Image Inpainting}& RePaint (CVPR'22) \cite{lugmayr2022repaint}& Diffusion& $512^2$& 54& CelebA-HQ \cite{liu2015deep} \\
      & MAT (CVPR'22) \cite{li2022mat}& GAN& $512^2$& 616& CelebA-HQ \cite{liu2015deep}, Places \cite{zhou2017places}\\
    \midrule
     \multirow{3}{*}{Image Colorization}& PDNLA-Net (TIP'23) \cite{wang2023unsupervised}& GAN& $512^2$& 191& ADE20K \cite{zhou2017scene} \\
      & DDColor (ICCV'23) \cite{kang2022ddcolor}& VAE& $\ge512^2$& 326& Imagenet \cite{russakovsky2015imagenet} \\
      & DISCO (SIGGRAPH'22) \cite{xia2022disentangled}& Diffusion& $512^2$& 289& COCO \cite{lin2014microsoft} \\
    \midrule
     \multirow{3}{*}{Image Editing}& DragGAN (SIGGRAPH'23) \cite{pan2023drag}& GAN& $512^2$& 203& DeepFashion \cite{liu2016deepfashion}  \\
      & InstructPix2Pix (CVPR'23) \cite{brooks2023instructpix2pix}& Diffusion& $\ge512^2$& 105& ADE20K \cite{zhou2017scene}, Landscape \cite{chen2018cartoongan} \\
      & MagicBrush (arXiv'23) \cite{zhang2023magicbrush}& GAN+Diffusion& $1024^2$& 519& COCO \cite{lin2014microsoft} \\
    \bottomrule[1.25pt]
  \end{tabular}}
  \label{appendix_model_diversity}
  \vspace{-1em}
\end{table*}

\subsection{Collecting AI-generated Images}
\noindent
{\bf Detailed Information of Text-to-Image Models.} 
For text-to-image models, we first choose 20 hot keywords from the \textit{PNGIMG} website\footnote{\url{https://pngimg.com}} and then create 10 prompts using GPT3.5 for each keyword with the following query:

\begin{quotation}
\noindent
\textit{I want you to act as a prompt generator for the text-to-image program. Your job is to provide detailed, accurate, and real descriptions that do exist in the real world and will inspire unique and true-life images from the AI. Now you can provide 10 detailed, accurate, and real descriptions (within 35 words) according to the keyword:} \texttt{<keyword>}.

\noindent
\textit{Keywords: nature, festival, food, animals, flower, people, space, travel, book, vehicles, artifacts, fruits, clothing, object, sport, electronics, transportation, architecture, drinks, human face.}
\end{quotation}
At times, the generated prompts are too obscure or too similar, so we manually examine the text compliance and restart the prompt generation process iteratively. We also randomly sample 100 captions from the COCO \cite{lin2014microsoft} dataset and choose 128 proper prompts from the commonly used DrawBench \cite{saharia2022photorealistic} by excluding inappropriate categories (\eg \textit{conflicting}, \textit{misspellings}, and \textit{rare words}).
Concretely, we select mainstream text-to-image models including Stable Diffusion v1.5\footnote{\url{https://huggingface.co/runwayml/stable-diffusion-v1-5}} \cite{rombach2022high}, Stable Diffusion v2.1\footnote{\url{https://huggingface.co/stabilityai/stable-diffusion-2-1}} \cite{rombach2022high}, Openjourney\footnote{\url{https://openjourney.art/}} \cite{midj,rombach2022high}, Dreamlike\footnote{\url{https://huggingface.co/dreamlike-art/dreamlike-photoreal-2.0}} \cite{dreamlike,rombach2022high}, and Realistic Vision v1.4\footnote{\url{https://huggingface.co/SG161222/Realistic_Vision_V1.4}} \cite{rombach2022high}, to generate photorealistic images of the aforementioned prompts.
Note that in real-world scenarios, people aim to use AI to obtain high-quality images without visual defects. Thus, we intentionally avoid using specialized prompt suffixes such as \textit{8K}, \textit{HDR}, \textit{photographic}, to simulate scenarios where naturalness issue occurs.

\noindent
{\bf Detailed Information of Image Translation Models.} We choose five up-to-date image translation models including RABIT \cite{zhan2022bi}, DiffuseIT \cite{kwon2022diffusion}, StyleCLIP \cite{patashnik2021styleclip}, MATEBIT \cite{jiang2023masked}, and CoCosNet \cite{zhang2020cross}, to investigate the manifestation of naturalness issues in image translation tasks. For RABIT, MATEBIT, and CoCosNet, we take different conditional inputs (\eg, edge map, semantic map) and exemplars provided by ADE20K \cite{zhou2017scene} and CelebA-HQ \cite{liu2015deep} datasets to generate translation results. For DiffuseIT, we use the pre-trained diffusion model provided by the authors and adopt both text-guided and image-guided strategies to achieve image translation, where target and source images are sampled from Landscape \cite{chen2018cartoongan} dataset. Conditional prompt keywords such as \textit{beach}, \textit{snow}, \textit{desert}, \textit{sea}, \textit{mountain}, \textit{cloud}, and \textit{grassfield}, are applied to improve the diversity of outputs. For StyleCLIP, we perform attribute changes including facial expression, hairstyle, skin color, and makeup, on the portraits of celebrities from FFHQ \cite{karras2019style} dataset using the pre-trained StyleGAN2 \cite{karras2020analyzing}.

\noindent
{\bf Detailed Information of Image Inpainting Models.} 
Image inpainting, also known as image completion, aims at filling missing regions reasonably within an image while maintaining harmonization with the rest of the image. 
Inpainting approaches thus require strong generative capabilities, otherwise, it can lead to poor results with severe naturalness distortion. 
Here, we randomly select input contents from CelebA-HQ \cite{liu2015deep} and Places \cite{zhou2017places} datasets and generate 54 and 616 images using RePaint \cite{lugmayr2022repaint} and MAT \cite{li2022mat}, respectively. Partial brushes, squares, or even masks with large missing areas are employed to enhance the diversity of generated contents.
We notice that both two models struggle in processing complex scenes with multiple objects due to the lack of sufficient semantic understanding. Most of the generated images have various degrees of artifacts and unreasonable layouts, which highlights the necessity to evaluate the naturalness of images.

\noindent
{\bf Detailed Information of Image Colorization Models.} 
Compared to the complete image generation task, image colorization aims to restore two missing color channels on the basis of grayscale images, which usually suffers from ambiguity and uncertainty. In other words, an object may accept multiple distinct colors while keeping the semantic consistency among pixels. Such characteristics predispose it to be a hard-hit area of naturalness problems. For this reason, we select three representative models including PDNLA-Net \cite{wang2023unsupervised}, DDColor \cite{kang2022ddcolor}, and DISCO \cite{xia2022disentangled}, to reflect this phenomenon. Specifically, we extract source images from ADE20K \cite{zhou2017scene}, Imagenet \cite{russakovsky2015imagenet}, and COCO \cite{lin2014microsoft} datasets and manually convert them to grayscale as input.

% Since the generative models reveal the powerful performance and potential in image generation

\begin{figure*}[t]
  \centering
   \includegraphics[width=1\linewidth]{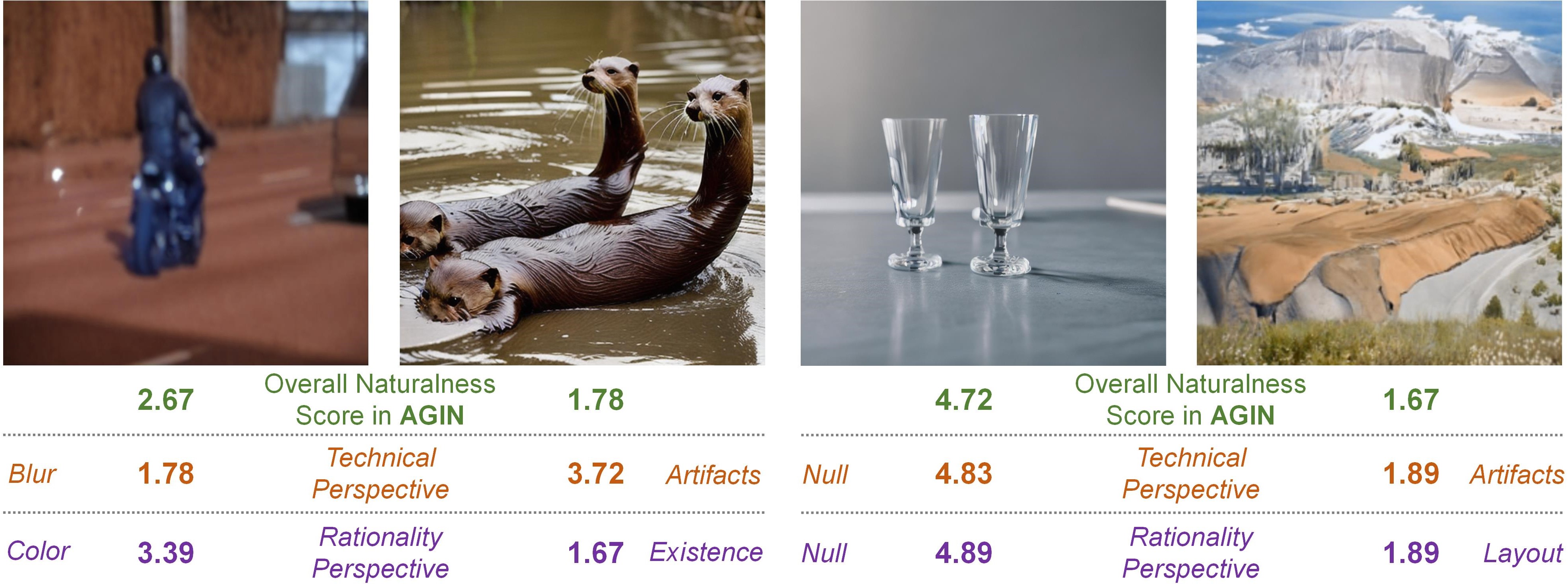}
   \caption{{\bf Four extended examples in AGIN that \textit{technical} and \textit{rationality} perspectives exhibit different effects on naturalness.} The two images on the left have different levels of technical and rationality scores, whereas the two images on the right have similar levels of technical and rationality scores. We also report the main factors with the highest proportion from the subject's selection for each perspective.}
   \label{extended}
   \vspace{-0.5cm}
\end{figure*}

\noindent
{\bf Detailed Information of Image Editing Models.}
There exists a huge demand for providing flexible and controllable image editing means in daily life, ranging from individual users to professional applications. Improper instruction may cause unnaturalness between the edited areas and surrounding contents.
Here, we choose the well-known text-guided image editing model (InstructPix2Pix \cite{brooks2023instructpix2pix}) and an interactive model (DragGAN \cite{pan2023drag}) to generate natural or unnatural images. For InstructPix2Pix, we apply text instructions, such as ``\textit{add sth to sth}", ``\textit{turn it into sth}", ``\textit{replace sth the with sth}", to edit images. The source images are from ADE20K \cite{zhou2017scene} and Landscape \cite{chen2018cartoongan} datasets.
For DragGAN, we choose images from the DeepFashion \cite{liu2016deepfashion} dataset and set up two anchor points with certain drag directions. Then, DragGAN will move the handle point to reach its corresponding target point, thus completing the image editing procedure.
Moreover, we extract 519 images from MagicBrush \cite{zhang2023magicbrush}, a large-scale, manually annotated dataset for instruction-guided real image editing. Since it already covers diverse editing scenarios, we do not re-process these images.

%------------------------------------------------------------

%----------------------------------------------------------------------

\begin{figure*}[t]
  \centering
   \includegraphics[width=1\linewidth]{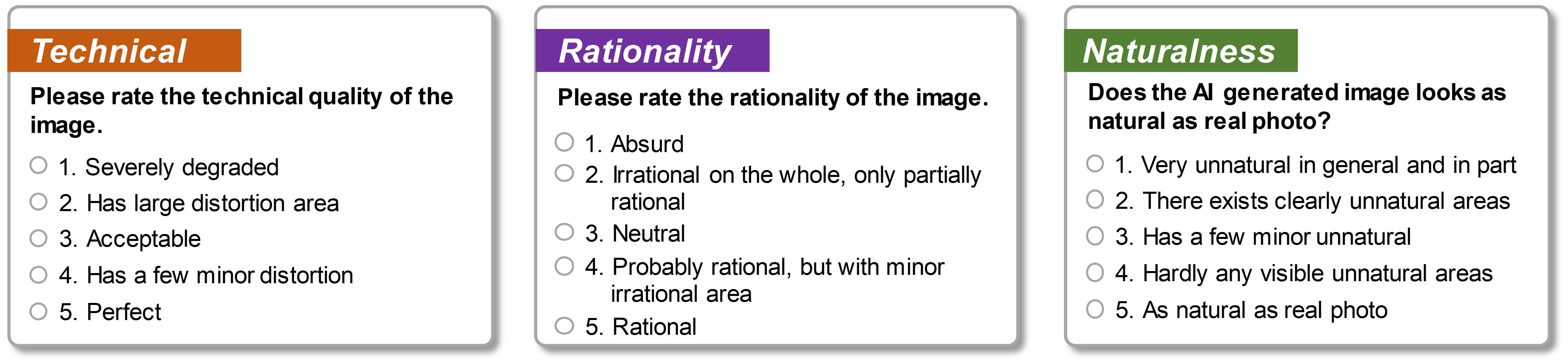}
   \caption{{\bf Question and labels of three candidate task designs.} Instead of using typical labels for a 5-point Likert scale, we elaborate questions and labels for the rating of technical quality, rationality, and naturalness.}
   \label{question}
   \vspace{-1em}
\end{figure*}

\begin{figure}[t]
  \centering
   \includegraphics[width=1\linewidth]{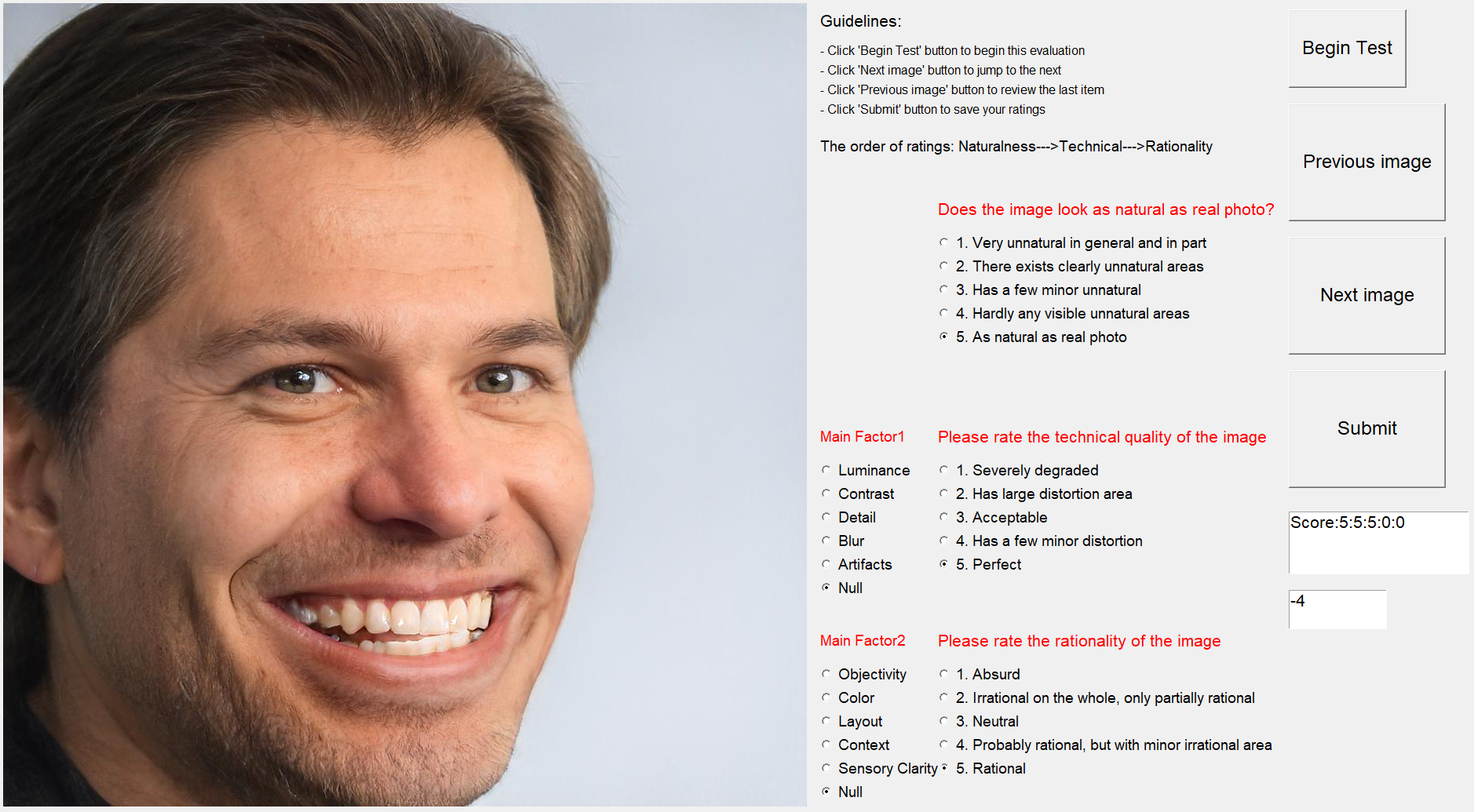}
   \caption{{\bf The interface for human evaluation interface.} Subjects are required to rate the technical quality, rationality, and naturalness of AGIs, and select the corresponding main factor through the radio buttons.}
   \label{interface}
   \vspace{-1.5em}
\end{figure}

\subsection{Detailed Information of Human Evaluation}
\label{extra_human}

\noindent
{\bf Laboratory Setup.}
Considering the viewing effect, a 27-inch Lenovo monitor with 2560×1440 resolution is used for image display. The viewing distance and optimal horizontal viewing angle
are set as 1.9 times the height of the display ($\approx70$cm) and [$31^{\circ}, 58^{\circ}$], respectively. Other settings such as ambient brightness, lighting, and background are configured according to the ITU-R BT.500 recommendation \cite{bt2002methodology}.

\noindent
{\bf Interface Design and Stimuli Presentation.}
As shown in Fig.  \ref{interface}, the interface layout is mainly composed of the left image display area and the right operation area, which allows viewers to browse the previous/next image and select the most appropriate options.
We adopt the single-stimulus procedure for naturalness assessment and require the participant to focus on and evaluate the overall naturalness, as well as the technical quality and rationality of the images. 
To avoid the interference of rating technical quality and rationality on naturalness collection, the evaluation of each image follows a 2-phase process. Firstly, an image is selected from AGIN database and participants are asked to evaluate the overall naturalness. Secondly, only after the naturalness evaluation is complete can participants move on to rate the technical quality and rationality of the image, as well as to select their respective main factors. Specifically, we design questions and labels for three different tasks as shown in Fig.  \ref{question}, which affects participants’ labeling behavior and enables us to obtain more accurate data. 
Besides, to maximize the rating efficiency, participants were asked to click 1-5 radio buttons directly instead of using the keyboard to enter.

\noindent
{\bf Formal Study.} Each participant was required to evaluate 6,049 images from three perspectives and to select two main factors, yielding a total of $6,049\times 30\times (3+2)=907,350$ ratings. 
Note that we shuffle and randomly divide all images into 15 sessions, each session except the 15th contains 400 images.
During the study, all subjects go through the \textit{spot check} in each session that they need to correctly rate the \textit{golden images} in a proper range (\textit{expert-set} rating $\pm 1$ for $>70\%$ images). Otherwise, the subject will be rejected for the next session.
Considering the large number of images, to reduce visual fatigue, there is a rest session with at least 15 minutes between two sessions.
To summarize, it took participants nearly 2 hours to finish one session, and all experiments were completed within a week.
Each participant was compensated \$16 for each session according to the current ethical standard \cite{silberman2018responsible}.

\noindent
{\bf Annotation Quality Assessment.} The reliability of results is of great importance while many studies did not report this entry. In this paper, we follow Otani \etal's \cite{otani2023toward} recommendation that uses the inter-annotator agreement (IAA) metric (Krippendorff’s $\alpha$ \cite{hayes2007answering}) to assess the quality of scoring. 
As a result, Krippendorff's $\alpha$ for technical quality, rationality, and naturalness ratings are 0.32, 0.33, and 0.37, respectively, indicating appropriate variations among annotators. Furthermore, we use SRCC as a criterion, calculating the correlation between each participant and MOSs, to judge whether an annotator is an outlier. We removed two participants with extremely low SRCC (0.1851 and 0.2839), resulting in an improvement on Krippendorff's $\alpha$ of 0.07 (from 0.32 to 0.39), 0.05 (from 0.33 to 0.38), and 0.04 (from 0.37 to 0.41) for technical quality, rationality, and naturalness scores, respectively.

\noindent
{\bf Mean Opinion Scores.} The mean opinion scores (MOSs) for each perspective are obtained by averaging cleaned ratings from different subjects. The final score is ranging from 1 to 5. Specifically, the minimum and maximum score values of the MOS\textsubscript{T} (technical perspective) are 1.39 and 5, and 1.17 and 5 for the MOS\textsubscript{R} (rationality perspective), 1.28 and 5 for the overall naturalness.

\noindent
{\bf Subjective Divergence between Perspectives.} In Fig. \ref{extended}, we further show four extended examples in AGIN, where the 1st image from left to right has worse \textit{technical quality} (more \textit{blurry}) while the 2nd image has significantly worse \textit{rationality} score due to the relatively \textit{nonexistent} contents. Such divergence (one has better technical quality, one has worse rationality) indicates that the overall naturalness scores not only depend on one aspect, which poses challenges to traditional IQA models supervised by a single MOS.
Meanwhile, the 3rd and 4th images illustrate a more common scenario, possessing similar scores in terms of technical and rationality aspects.

\begin{figure*}[t]
  \centering
   \includegraphics[width=1\linewidth]{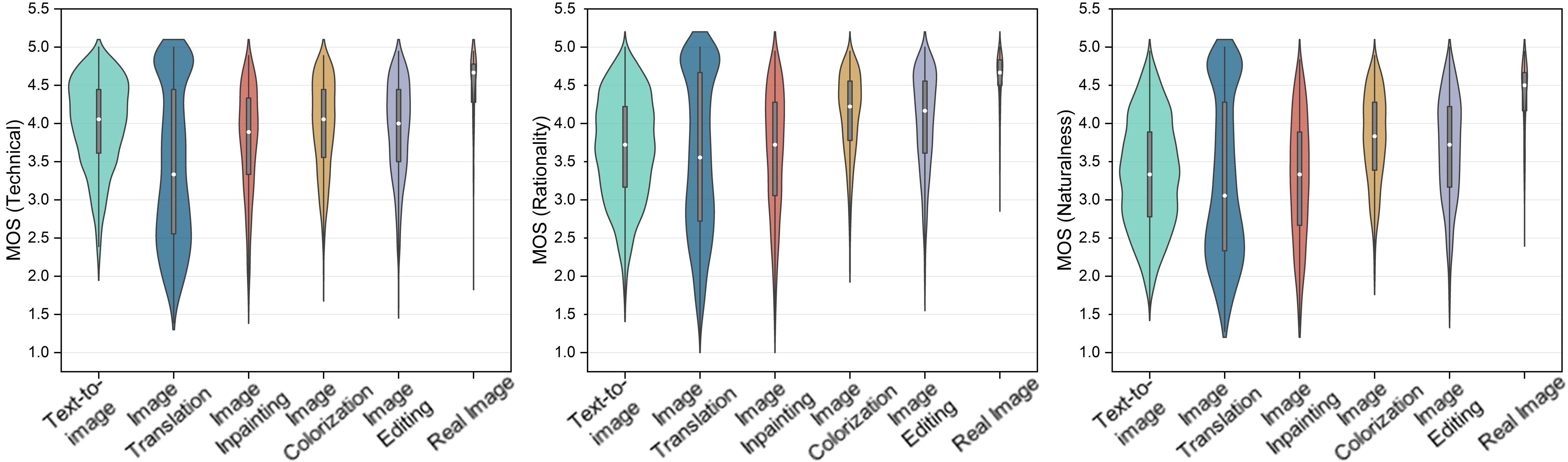}
   \caption{{\bf Visualization of all categories in AGIN.} From left to right are the violin plots of technical quality, rationality, and overall naturalness scores. For each box, the white circle indicates the median and the edges of the box represent the 25th and 75th percentiles. }
   \label{violin_all}
   \vspace{-1.5em}
\end{figure*}

\begin{figure*}[t]
  \centering
   \includegraphics[width=1\linewidth]{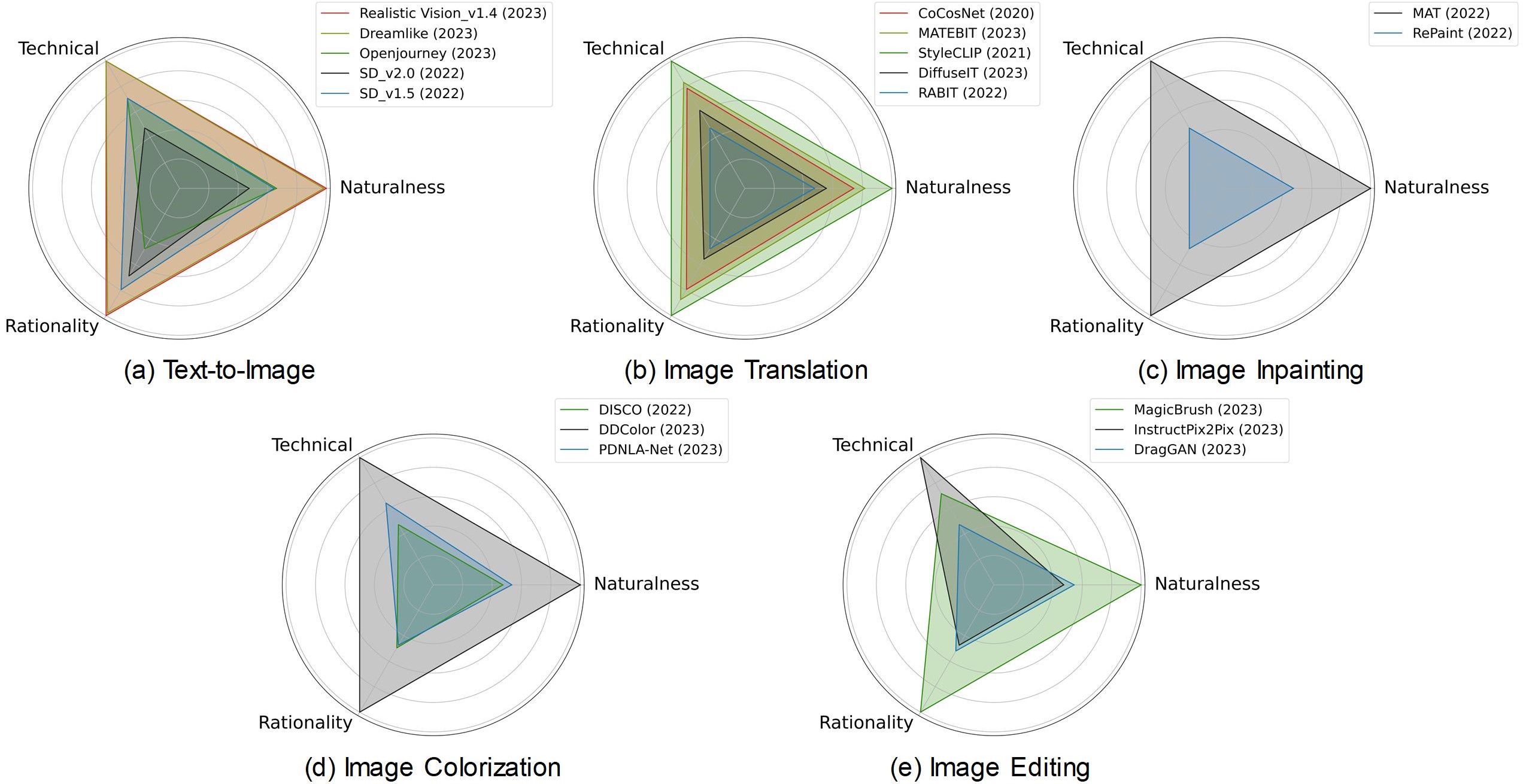}
   \caption{More comparisons of image generation models. We visualize the {\bf averaged MOS} for technical, rationality, and overall naturalness perspectives.}
   \label{mos_radar}
   \vspace{-1em}
\end{figure*}

\begin{figure*}[t]
  \centering
   \includegraphics[width=1\linewidth]{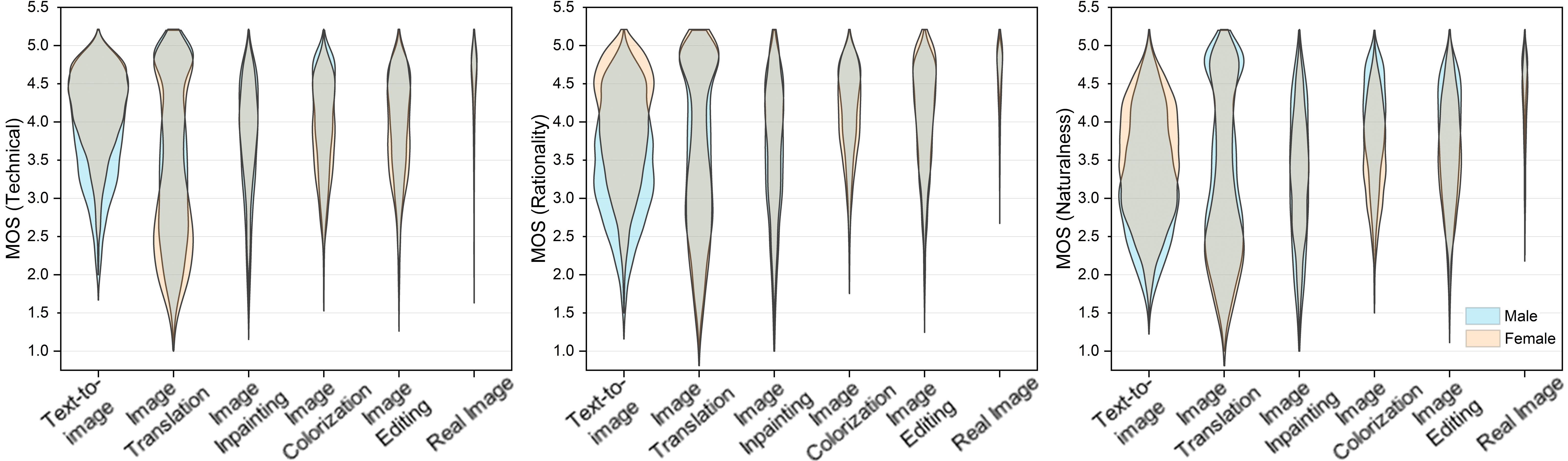}
   \caption{Score distributions for {\bf male and female subjects} with different categories.}
   \label{violin_mafe}
   \vspace{-1em}
\end{figure*}

\subsection{Detailed Score Distribution}
\noindent
{\bf Detailed score distribution of different categories.} As shown in Fig. \ref{violin_all}, we visualize the detailed score distribution of different categories for all participants. It can be observed that the real images in AGIN have relatively high technical, rationality, and naturalness scores, which is consistent with our intention of controlling the quality of the rating results. Besides, AGIs from the image translation task have the widest distribution and lowest average scores for technical, rationality, and naturalness. 
We can roughly observe from Fig. \ref{mos_radar} that the naturalness and its two perspectives of AGIs have improved over the years.
In general, regardless of the number of images, all categories have a relatively balanced score distribution in terms of technical quality, rationality, and naturalness.

\noindent
{\bf Detailed score distribution of different categories in terms of gender.} We also visualize the detailed score distribution of different categories for males and females in Fig. \ref{violin_mafe}. We can observe that the score distribution of different categories for males and females is basically consistent except for the category of \textit{text-to-image}. 
We speculate that people may have cognitive differences for familiar or unfamiliar objects, where women are more sensitive than men.
%Besides, the MOS\textsubscript{R} females may be more sensitive to color
%-----------------------------------------------------------------
\begin{figure}[t]
  \centering
   \includegraphics[width=1\linewidth]{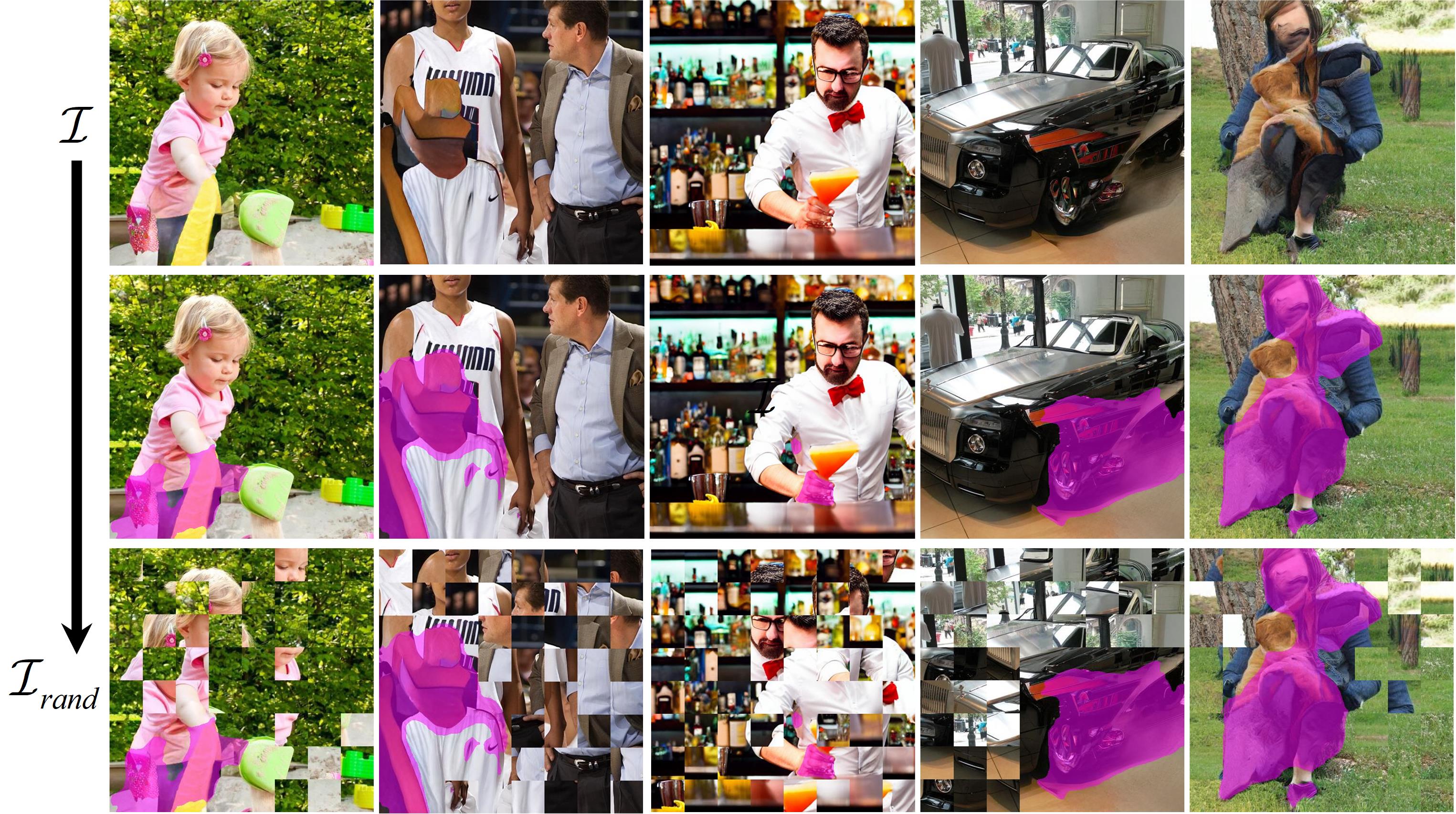}
   \caption{{\bf Visualization of perceptual artifacts-guided patch partition.} The first row contains generated images from AGIN. The second row exhibits the prediction results of perceptual artifacts using the pre-trained model in \cite{zhang2023perceptual}. In the last row, we show the resulted $\mathcal{I}_{rand}$ after perceptual artifacts-guided patch partition.}
   \label{irand}
   \vspace{-1em}
\end{figure}

%-----------------------------------------------------------------
\begin{figure}[t]
  \centering
   \includegraphics[width=1\linewidth]{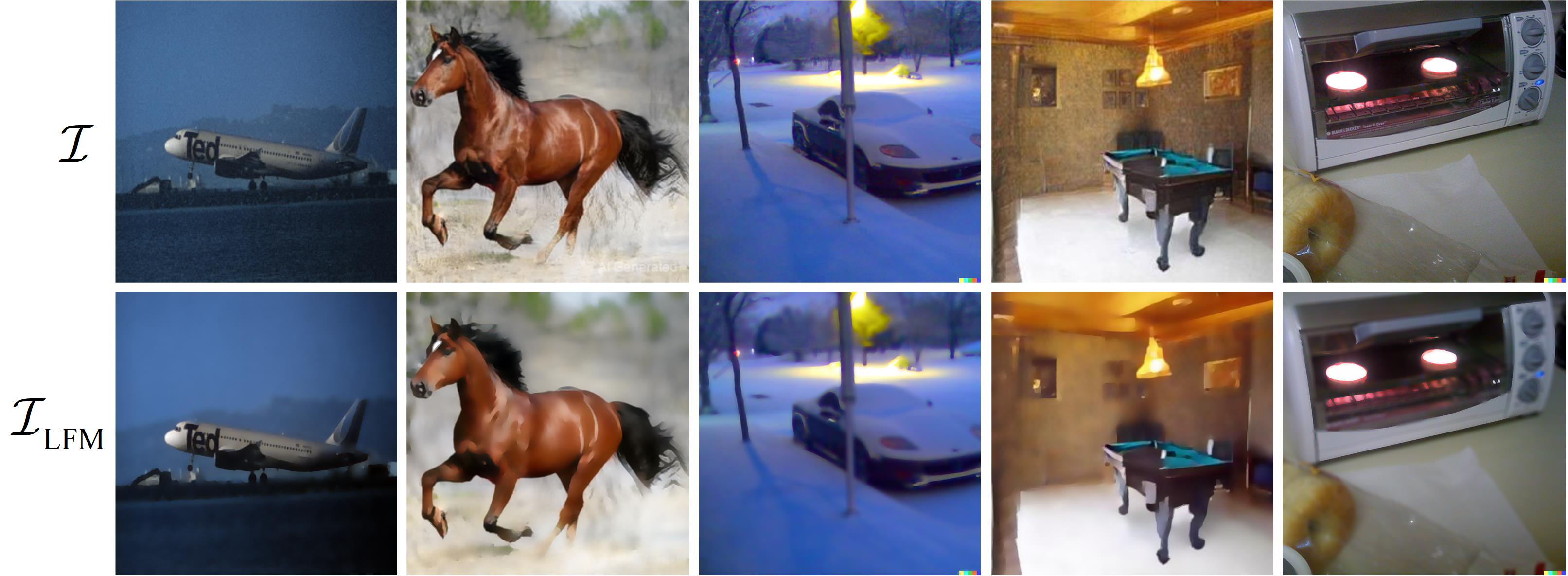}
   \caption{Comparison of the original image and its corresponding low-frequency map.}
   \label{LFM}
\end{figure}
%-----------------------------------------------------------------

\section{Detailed Structure of the JOINT}
\noindent
In this section, we introduce some definitions and necessary notations that will be used in JOINT. Examples of the perceptual artifacts-guided patch partition in the technical prior branch are illustrated in Fig. \ref{irand}.
\subsection{Low-Frequency Map}
\noindent
To maintain the principal semantic information of the image while better reducing the impact of partial technical distortions, we utilize the piece-wise smooth image approximation algorithm \cite{Bar2006SemiblindIR} to generate the low-frequency map by minimizing: 
\begin{equation}
\begin{split}
\mathcal{F} = \frac{1}{2}\int_\Omega  {{\left( {\mathcal{I} - {\mathcal{I}_{{\mathrm{LFM}}}}} \right)}^2}dP &+ \mu \int_{\Omega \backslash E} {{\left| {\nabla {\mathcal{I}_{{\mathrm{LFM}}}}} \right|}^2}dP \\
&+ \nu \int_E {d\sigma},  
\end{split}
\end{equation}
where $\Omega$ and $E$ denote the image domain and edge set, respectively. $P$ indicates the pixel and $\int_E {d\sigma}$ represents the total edge length. The coefficients $\mu$ and $\nu$ are positive regularization constants. An example of low-frequency maps is shown in Fig. \ref{LFM}. We can observe that the LFM filters out some technical distortions but still preserves similar semantic information to the original image.

\subsection{Definition of Wasserstein Distance}
\noindent
Given two multidimensional random variables $P$ and $Q$ with their distributions denoted as $\mathcal{X}$ and $\mathcal{Y}$, respectively, the $l$-Wasserstein distance between them is defined as
\begin{equation}
{W_l}(P,Q): = {\left( {\mathop {\inf }\limits_{\gamma  \in \mathcal{J}\left( {\mathcal{X},\mathcal{Y}} \right)} \int {{{\left\| {p - q} \right\|}_l}d\gamma (p,q)} } \right)^{1/l}},
\label{wsd}
\end{equation}
where $p$ and $q$ are the masses of $P$ and $Q$. $\gamma  \in \mathcal{J}\left( {\mathcal{X},\mathcal{Y}}\right)$ is the joint distribution of $(P,Q)$. $l$ is the order of the $l$-norm.
Additionally, for one-dimensional probability measures, the Eq. \ref{wsd} is closed-form \cite{cazelles2020wasserstein} and boils down to
\begin{equation}
{W_l}(P,Q) = {\left( {\int_0^1 {{{\left| {F_p^ - (t) - F_q^ - (t)} \right|}^p}dt} } \right)^{1/l}},
\end{equation}
where $F_p^-$ represents the inverse cumulative function of $P$. $t$ is the implicit variable that is used to integral $F_p^ -(\cdot)$ and $F_q^ - (\cdot)$ from 0 to 1.

\subsection{Training Objective}
Here, we discuss the concrete designs of the combination of the standard mean square error 
loss $\mathcal{L}_{\mathrm{MSE}}$ and Spearman Rank-order Correlation Coefficient (SRCC) loss $\mathcal{L}_{\mathrm{SRCC}}$ as follows:
\begin{align}
    {\mathcal{L}_{{\mathrm{MSE}}}} &= \frac{1}{N}\sum\nolimits_{n = 1}^N {\left\| {{y_n} - {{\hat y}_n}} \right\|} _2^2, \\
    {\mathcal{L}_{{\mathrm{SRCC}}}} &= 1 - \frac{{\sum\nolimits_n {({v_n} - \bar v)({p_n} - \bar p)} }}{{\sqrt {\sum\nolimits_n {{{({v_n} - \bar v)}^2}\sum\nolimits_n {{{({p_n} - \bar p)}^2}} } } }}, \\
    {\mathcal{L}_{\mathrm{C}}} &= {\mathcal{L}_{{\mathrm{MSE}}}} + {\mathcal{L}_{{\mathrm{SRCC}}}},
\end{align}
where the SRCC is defined in the form of the Pearson linear correlation coefficient (PLCC) between ranks \cite{blondel2020fast, Li2022BlindlyAQ}. $v_n$ and $p_n$ denote the rank of the ground truth $y_n$ and the rank of predicted score ${\hat y_n}$, respectively.

\section{More Experimental Details}
\subsection{Implementation Details}
\noindent
We initialize all baselines using their own implementations and hyperparameters. In the rationality branch, all images are calculated at size $224\times 224$ so as to satisfy the requirement of ResNet50. Besides, for penalty constraint $\mathcal{L}_{\mathrm{WSD}}$, we set $l=2$ as in \cite{Liao2022DeepWSDPD}, making the quality measure more sensitive to outliers. 
The $N$ in $\mathcal{L}_{\mathrm{WSD}}$ is $5$, corresponding to five stages in Resnet50 backbone. 
Before training, we randomly split the training, validation, and testing set into 7:1:2. There is no overlap of the same image content between each set. We repeat the split process 5 times and record the average performance as the final experimental results.

\subsection{Evaluation Metrics}
\noindent
We adopt the widely used metrics in IQA literature \cite{zhai2020perceptual}: Spearman rank-order correlation coefficient (SRCC) and Pearson linear correlation coefficient (PLCC), as our evaluation criteria. SRCC quantifies the extent to which the ranks of two variables are related, which ranges from -1 to 1. Given $N$ distorted images, SRCC is computed as:
\begin{equation}
SRCC = 1 - \frac{{6\sum\nolimits_{n = 1}^N {{{({v_n} - {p_n})}^2}} }}{{N({N^2} - 1)}},
\end{equation}
where $v_n$ and $p_n$ denote the rank of the ground truth $y_n$ and the rank of predicted score ${\hat y_n}$ respectively. The higher the SRCC, the higher the monotonic correlation between ground truth and predicted score.
Similarly, PLCC measures the linear correlation between predicted scores and ground truth scores, which can be formulated as:
\begin{equation}
PLCC = \frac{{\sum\nolimits_{n = 1}^N {({y_n} - \bar y)({{\hat y}_n} - \bar {\hat y})} }}{{\sqrt {\sum\nolimits_{n = 1}^N {{{({y_n} - \bar y)}^2}} } \sqrt {\sum\nolimits_{n = 1}^N {{{({{\hat y}_n} - \bar {\hat y})}^2}} } }},
\end{equation}
where $\bar y$ and $\bar {\hat y}$ are the mean of ground truth and predicted score respectively.
%-----------------------------------------------------------------------
\begin{figure}[t]
  \centering
   \includegraphics[width=1\linewidth]{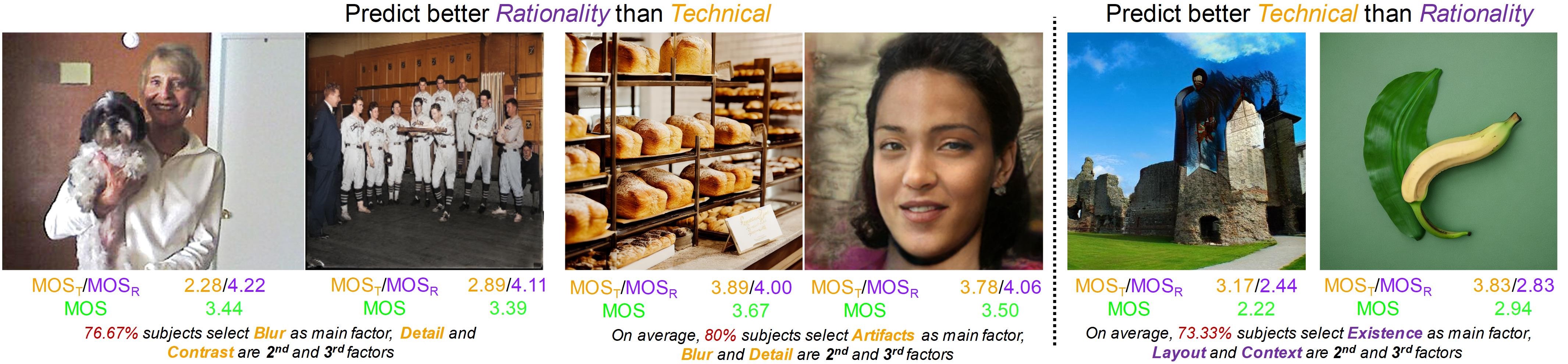}
   \caption{Success case of JOINT.}
   \label{success}
   \vspace{-1em}
\end{figure}

\begin{figure}[t]
  \centering
   \includegraphics[width=1\linewidth]{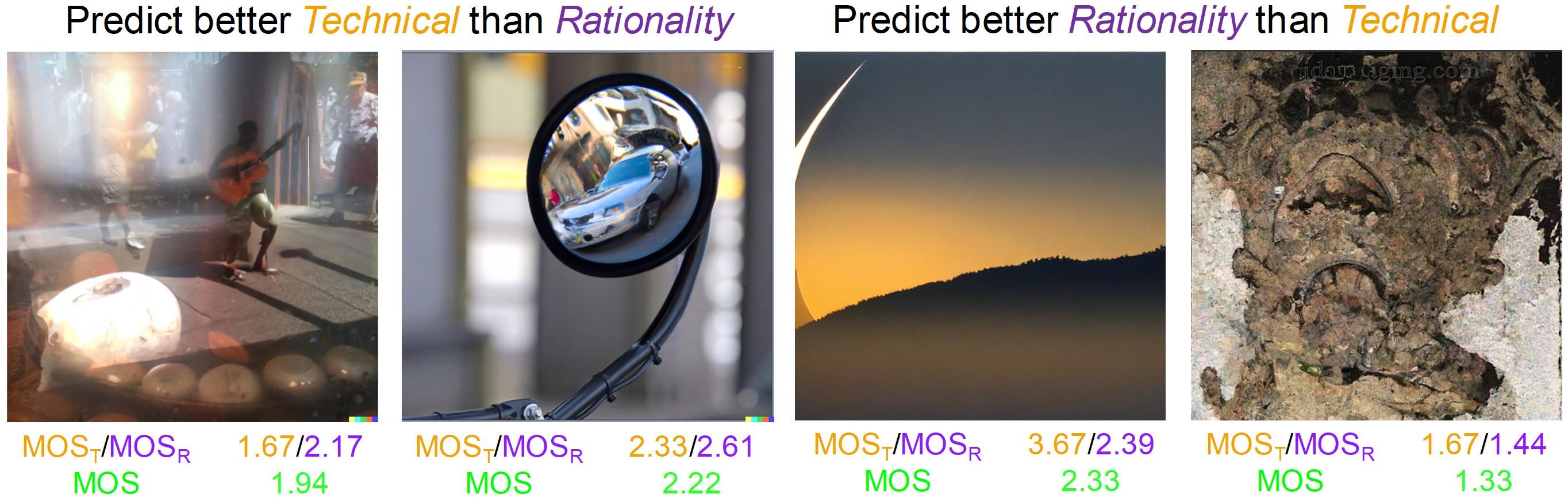}
   \caption{Failure case of JOINT.}
   \label{fail}
   \vspace{-2em}
\end{figure}

%-----------------------------------------------------------------------

\subsection{Extended Qualitative Results}
\noindent
{\bf Success Cases.}
We visualized several successful cases (\textit{i.e.}, when the predicted results are consistent with subjective annotators in each dimension) for the proposed JOINT in Fig. \ref{success}. Specifically, we can observe that our JOINT is sensitive to global technical distortions such as blur, lack of detail, and contrast (1st and 2nd images in Fig. \ref{success}). Moreover, the perceptual artifacts-guided patch partition strategy endows JOINT with the ability to measure the severity of local artifacts (3rd and 4th images in Fig. \ref{success}). On the contrary, the rationality perceiving branch is not sensitive to technical distortion since we add deep feature regularization with filtered low-frequency information, and it is very sensitive to the rationality of contents and can recognize the very unusual compositions and objects in images (the left images in Fig. \ref{success}).
All these cases further demonstrate the effectiveness of the JOINT in jointly learning from both perspectives in image naturalness assessment.

\noindent
{\bf Failure Cases.}
As for the failure cases, we notice that most of them are difficult to recognize (\textit{poor sensory clarity}, the \textit{right} two in Fig. \ref{fail}) or are extremely blurry with a small part of clear areas (\textit{mirror reflections}, the \textit{left} two in Fig. \ref{fail}). Such cases are more frequent in photography that photographers gain prominence over the target by blurring the background, yet the model can experience challenges in predicting technical quality and rationality, resulting in a failure in naturalness evaluation.

%\bibliographystyle{splncs04}
%\bibliography{main}

\begin{thebibliography}{100}
\providecommand{\url}[1]{\texttt{#1}}
\providecommand{\urlprefix}{URL }
\providecommand{\doi}[1]{https://doi.org/#1}

\bibitem{agustsson2017ntire}
Agustsson, E., Timofte, R.: Ntire 2017 challenge on single image super-resolution: Dataset and study. In: Proceedings of the IEEE conference on computer vision and pattern recognition workshops. pp. 126--135 (2017)

\bibitem{Bar2006SemiblindIR}
Bar, L., Sochen, N., Kiryati, N.: Semi-blind image restoration via mumford-shah regularization. IEEE Transactions on Image Processing  \textbf{15},  483--493 (2006)

\bibitem{beghdadi2022benchmarking}
Beghdadi, A., Mallem, M., Beji, L.: Benchmarking performance of object detection under image distortions in an uncontrolled environment. In: 2022 IEEE International Conference on Image Processing (ICIP). pp. 2071--2075. IEEE (2022)

\bibitem{binkowski2018demystifying}
Bi{\'n}kowski, M., Sutherland, D.J., Arbel, M., Gretton, A.: Demystifying mmd gans. arXiv preprint arXiv:1801.01401  (2018)

\bibitem{blondel2020fast}
Blondel, M., Teboul, O., Berthet, Q., Djolonga, J.: Fast differentiable sorting and ranking. In: International Conference on Machine Learning. pp. 950--959. PMLR (2020)

\bibitem{brooks2023instructpix2pix}
Brooks, T., Holynski, A., Efros, A.A.: Instructpix2pix: Learning to follow image editing instructions. In: Proceedings of the IEEE/CVF Conference on Computer Vision and Pattern Recognition. pp. 18392--18402 (2023)

\bibitem{bt2002methodology}
BT, R.: Methodology for the subjective assessment of the quality of television pictures. International Telecommunication Union  \textbf{4} (2002)

\bibitem{cadfk2005naturalness}
Cadfk, M., Slav{\'\i}k, P.: The naturalness of reproduced high dynamic range images. In: Ninth International Conference on Information Visualisation (IV'05). pp. 920--925. IEEE (2005)

\bibitem{cao2023comprehensive}
Cao, Y., Li, S., Liu, Y., Yan, Z., Dai, Y., Yu, P.S., Sun, L.: A comprehensive survey of ai-generated content (aigc): A history of generative ai from gan to chatgpt. arXiv preprint arXiv:2303.04226  (2023)

\bibitem{cazelles2020wasserstein}
Cazelles, E., Robert, A., Tobar, F.: The wasserstein-fourier distance for stationary time series. IEEE Transactions on Signal Processing  \textbf{69},  709--721 (2020)

\bibitem{chen2018cartoongan}
Chen, Y., Lai, Y.K., Liu, Y.J.: Cartoongan: Generative adversarial networks for photo cartoonization. In: Proceedings of the IEEE conference on computer vision and pattern recognition. pp. 9465--9474 (2018)

\bibitem{chen2023x}
Chen, Y.: X-iqe: explainable image quality evaluation for text-to-image generation with visual large language models. arXiv preprint arXiv:2305.10843  (2023)

\bibitem{chen2023quantifying}
Chen, Y., Akhtar, N., Haldar, N.A.H., Mian, A.: On quantifying and improving realism of images generated with diffusion. arXiv preprint arXiv:2309.14756  (2023)

\bibitem{choi2009investigation}
Choi, S.Y., Luo, M., Pointer, M., Rhodes, P.: Investigation of large display color image appearance--iii: Modeling image naturalness. Journal of Imaging Science and Technology  \textbf{53}(3),  31104--1 (2009)

\bibitem{ciancio2010no}
Ciancio, A., da~Silva, E.A., Said, A., Samadani, R., Obrador, P., et~al.: No-reference blur assessment of digital pictures based on multifeature classifiers. IEEE Transactions on image processing  \textbf{20}(1),  64--75 (2010)

\bibitem{cristani2013unveiling}
Cristani, M., Vinciarelli, A., Segalin, C., Perina, A.: Unveiling the multimedia unconscious: Implicit cognitive processes and multimedia content analysis. In: Proceedings of the 21st ACM international conference on Multimedia. pp. 213--222 (2013)

\bibitem{dreamlike}
Dreamlike.art: {https://dreamlike.art}, \url{https://dreamlike.art}, 2023

\bibitem{fang2020perceptual}
Fang, Y., Zhu, H., Zeng, Y., Ma, K., Wang, Z.: Perceptual quality assessment of smartphone photography. In: Proceedings of the IEEE/CVF Conference on Computer Vision and Pattern Recognition. pp. 3677--3686 (2020)

\bibitem{ghadiyaram2015massive}
Ghadiyaram, D., Bovik, A.C.: Massive online crowdsourced study of subjective and objective picture quality. IEEE Transactions on Image Processing  \textbf{25}(1),  372--387 (2015)

\bibitem{goodale1992separate}
Goodale, M.A., Milner, A.D.: Separate visual pathways for perception and action. Trends in neurosciences  \textbf{15}(1),  20--25 (1992)

\bibitem{gu2022ntire}
Gu, J., Cai, H., Dong, C., Ren, J.S., Timofte, R., Gong, Y., Lao, S., Shi, S., Wang, J., Yang, S., et~al.: Ntire 2022 challenge on perceptual image quality assessment. In: Proceedings of the IEEE/CVF Conference on Computer Vision and Pattern Recognition. pp. 951--967 (2022)

\bibitem{gu2016blind}
Gu, K., Wang, S., Zhai, G., Ma, S., Yang, X., Lin, W., Zhang, W., Gao, W.: Blind quality assessment of tone-mapped images via analysis of information, naturalness, and structure. IEEE Transactions on Multimedia  \textbf{18}(3),  432--443 (2016)

\bibitem{guo2021underwater}
Guo, P., He, L., Liu, S., Zeng, D., Liu, H.: Underwater image quality assessment: Subjective and objective methods. IEEE Transactions on Multimedia  \textbf{24},  1980--1989 (2021)

\bibitem{hayes2007answering}
Hayes, A.F., Krippendorff, K.: Answering the call for a standard reliability measure for coding data. Communication methods and measures  \textbf{1}(1),  77--89 (2007)

\bibitem{he2016deep}
He, K., Zhang, X., Ren, S., Sun, J.: Deep residual learning for image recognition. In: Proceedings of the IEEE conference on computer vision and pattern recognition. pp. 770--778 (2016)

\bibitem{he2023thinking}
He, S., Ming, A., Li, Y., Sun, J., Zheng, S., Ma, H.: Thinking image color aesthetics assessment: Models, datasets and benchmarks. In: Proceedings of the IEEE/CVF International Conference on Computer Vision. pp. 21838--21847 (2023)

\bibitem{he2022rethinking}
He, S., Zhang, Y., Xie, R., Jiang, D., Ming, A.: Rethinking image aesthetics assessment: Models, datasets and benchmarks. In: Proceedings of the Thirty-First International Joint Conference on Artificial Intelligence, IJCAI-22. pp. 942--948 (2022)

\bibitem{hessel2021clipscore}
Hessel, J., Holtzman, A., Forbes, M., Bras, R.L., Choi, Y.: Clipscore: A reference-free evaluation metric for image captioning. arXiv preprint arXiv:2104.08718  (2021)

\bibitem{heusel2017gans}
Heusel, M., Ramsauer, H., Unterthiner, T., Nessler, B., Hochreiter, S.: Gans trained by a two time-scale update rule converge to a local nash equilibrium. Advances in neural information processing systems  \textbf{30} (2017)

\bibitem{hosu2020koniq}
Hosu, V., Lin, H., Sziranyi, T., Saupe, D.: Koniq-10k: An ecologically valid database for deep learning of blind image quality assessment. IEEE Transactions on Image Processing  \textbf{29},  4041--4056 (2020)

\bibitem{ingle1973two}
Ingle, D.: Two visual systems in the frog. Science  \textbf{181}(4104),  1053--1055 (1973)

\bibitem{jiang2023masked}
Jiang, C., Gao, F., Ma, B., Lin, Y., Wang, N., Xu, G.: Masked and adaptive transformer for exemplar based image translation. In: Proceedings of the IEEE/CVF Conference on Computer Vision and Pattern Recognition. pp. 22418--22427 (2023)

\bibitem{jinjin2020pipal}
Jinjin, G., Haoming, C., Haoyu, C., Xiaoxing, Y., Ren, J.S., Chao, D.: Pipal: a large-scale image quality assessment dataset for perceptual image restoration. In: Computer Vision--ECCV 2020: 16th European Conference, Glasgow, UK, August 23--28, 2020, Proceedings, Part XI 16. pp. 633--651. Springer (2020)

\bibitem{kang2022ddcolor}
Kang, X., Yang, T., Ouyang, W., Ren, P., Li, L., Xie, X.: Ddcolor: Towards photo-realistic and semantic-aware image colorization via dual decoders. arXiv preprint arXiv:2212.11613  (2022)

\bibitem{karras2019style}
Karras, T., Laine, S., Aila, T.: A style-based generator architecture for generative adversarial networks. In: Proceedings of the IEEE/CVF conference on computer vision and pattern recognition. pp. 4401--4410 (2019)

\bibitem{karras2020analyzing}
Karras, T., Laine, S., Aittala, M., Hellsten, J., Lehtinen, J., Aila, T.: Analyzing and improving the image quality of stylegan. In: Proceedings of the IEEE/CVF conference on computer vision and pattern recognition. pp. 8110--8119 (2020)

\bibitem{ke2021musiq}
Ke, J., Wang, Q., Wang, Y., Milanfar, P., Yang, F.: Musiq: Multi-scale image quality transformer. In: Proceedings of the IEEE/CVF International Conference on Computer Vision. pp. 5148--5157 (2021)

\bibitem{kingma2014adam}
Kingma, D.P., Ba, J.: Adam: A method for stochastic optimization. arXiv preprint arXiv:1412.6980  (2014)

\bibitem{kirstain2023pick}
Kirstain, Y., Polyak, A., Singer, U., Matiana, S., Penna, J., Levy, O.: Pick-a-pic: An open dataset of user preferences for text-to-image generation. arXiv preprint arXiv:2305.01569  (2023)

\bibitem{kwon2022diffusion}
Kwon, G., Ye, J.C.: Diffusion-based image translation using disentangled style and content representation. In: ICLR (2023)

\bibitem{kynkaanniemi2019improved}
Kynk{\"a}{\"a}nniemi, T., Karras, T., Laine, S., Lehtinen, J., Aila, T.: Improved precision and recall metric for assessing generative models. Advances in Neural Information Processing Systems  \textbf{32} (2019)

\bibitem{larson2010most}
Larson, E.C., Chandler, D.M.: Most apparent distortion: full-reference image quality assessment and the role of strategy. Journal of electronic imaging  \textbf{19}(1),  011006--011006 (2010)

\bibitem{le2020study}
Le, Q.T., Ladret, P., Nguyen, H.T., Caplier, A.: Study of naturalness in tone-mapped images. Computer Vision and Image Understanding  \textbf{196},  102971 (2020)

\bibitem{li2023image}
Li, B., Lu, Y., Pang, W., Xu, H.: Image colorization using cyclegan with semantic and spatial rationality. Multimedia Tools and Applications pp. 1--15 (2023)

\bibitem{Li2022BlindlyAQ}
Li, B., Zhang, W., Tian, M., Zhai, G., Wang, X.: Blindly assess quality of in-the-wild videos via quality-aware pre-training and motion perception. IEEE Transactions on Circuits and Systems for Video Technology  \textbf{32}(9),  5944--5958 (2022)

\bibitem{li2023agiqa}
Li, C., Zhang, Z., Wu, H., Sun, W., Min, X., Liu, X., Zhai, G., Lin, W.: Agiqa-3k: An open database for ai-generated image quality assessment. IEEE Transactions on Circuits and Systems for Video Technology  (2023). \doi{10.1109/TCSVT.2023.3319020}

\bibitem{li2020norm}
Li, D., Jiang, T., Jiang, M.: Norm-in-norm loss with faster convergence and better performance for image quality assessment. In: Proceedings of the 28th ACM International Conference on Multimedia. pp. 789--797 (2020)

\bibitem{li2020personality}
Li, L., Zhu, H., Zhao, S., Ding, G., Lin, W.: Personality-assisted multi-task learning for generic and personalized image aesthetics assessment. IEEE Transactions on Image Processing  \textbf{29},  3898--3910 (2020)

\bibitem{li2023towards}
Li, S., Zhang, S., Chen, G., Wang, D., Feng, P., Wang, J., Liu, A., Yi, X., Liu, X.: Towards benchmarking and assessing visual naturalness of physical world adversarial attacks. In: Proceedings of the IEEE/CVF Conference on Computer Vision and Pattern Recognition. pp. 12324--12333 (2023)

\bibitem{li2022mat}
Li, W., Lin, Z., Zhou, K., Qi, L., Wang, Y., Jia, J.: Mat: Mask-aware transformer for large hole image inpainting. In: Proceedings of the IEEE/CVF conference on computer vision and pattern recognition. pp. 10758--10768 (2022)

\bibitem{Liao2022DeepWSDPD}
Liao, X., Chen, B., Zhu, H., Wang, S., Zhou, M., Kwong, S.: Deepwsd: Projecting degradations in perceptual space to wasserstein distance in deep feature space. Proceedings of the 30th ACM International Conference on Multimedia  (2022)

\bibitem{lin2019kadid}
Lin, H., Hosu, V., Saupe, D.: Kadid-10k: A large-scale artificially distorted iqa database. In: 2019 Eleventh International Conference on Quality of Multimedia Experience (QoMEX). pp.~1--3. IEEE (2019)

\bibitem{lin2014microsoft}
Lin, T.Y., Maire, M., Belongie, S., Hays, J., Perona, P., Ramanan, D., Doll{\'a}r, P., Zitnick, C.L.: Microsoft coco: Common objects in context. In: Computer Vision--ECCV 2014: 13th European Conference, Zurich, Switzerland, September 6-12, 2014, Proceedings, Part V 13. pp. 740--755. Springer (2014)

\bibitem{liu2019unsupervised}
Liu, Y., Gu, K., Zhang, Y., Li, X., Zhai, G., Zhao, D., Gao, W.: Unsupervised blind image quality evaluation via statistical measurements of structure, naturalness, and perception. IEEE Transactions on Circuits and Systems for Video Technology  \textbf{30}(4),  929--943 (2019)

\bibitem{liu2021swin}
Liu, Z., Lin, Y., Cao, Y., Hu, H., Wei, Y., Zhang, Z., Lin, S., Guo, B.: Swin transformer: Hierarchical vision transformer using shifted windows. In: Proceedings of the IEEE/CVF international conference on computer vision. pp. 10012--10022 (2021)

\bibitem{liu2016deepfashion}
Liu, Z., Luo, P., Qiu, S., Wang, X., Tang, X.: Deepfashion: Powering robust clothes recognition and retrieval with rich annotations. In: Proceedings of the IEEE conference on computer vision and pattern recognition. pp. 1096--1104 (2016)

\bibitem{liu2015deep}
Liu, Z., Luo, P., Wang, X., Tang, X.: Deep learning face attributes in the wild. In: Proceedings of the IEEE international conference on computer vision. pp. 3730--3738 (2015)

\bibitem{lu2023seeing}
Lu, Z., Huang, D., Bai, L., Liu, X., Qu, J., Ouyang, W.: Seeing is not always believing: A quantitative study on human perception of ai-generated images. arXiv preprint arXiv:2304.13023  (2023)

\bibitem{lugmayr2022repaint}
Lugmayr, A., Danelljan, M., Romero, A., Yu, F., Timofte, R., Van~Gool, L.: Repaint: Inpainting using denoising diffusion probabilistic models. In: Proceedings of the IEEE/CVF Conference on Computer Vision and Pattern Recognition. pp. 11461--11471 (2022)

\bibitem{ma2016waterloo}
Ma, K., Duanmu, Z., Wu, Q., Wang, Z., Yong, H., Li, H., Zhang, L.: Waterloo exploration database: New challenges for image quality assessment models. IEEE Transactions on Image Processing  \textbf{26}(2),  1004--1016 (2016)

\bibitem{midj}
Midjourney: {https://www.midjourney.com}, \url{https://www.midjourney.com}, 2023

\bibitem{mittal2012no}
Mittal, A., Moorthy, A.K., Bovik, A.C.: No-reference image quality assessment in the spatial domain. IEEE Transactions on image processing  \textbf{21}(12),  4695--4708 (2012)

\bibitem{mittal2012making}
Mittal, A., Soundararajan, R., Bovik, A.C.: Making a “completely blind” image quality analyzer. IEEE Signal processing letters  \textbf{20}(3),  209--212 (2013)

\bibitem{murray2012ava}
Murray, N., Marchesotti, L., Perronnin, F.: Ava: A large-scale database for aesthetic visual analysis. In: 2012 IEEE conference on computer vision and pattern recognition. pp. 2408--2415. IEEE (2012)

\bibitem{nichol2021glide}
Nichol, A., Dhariwal, P., Ramesh, A., Shyam, P., Mishkin, P., McGrew, B., Sutskever, I., Chen, M.: Glide: Towards photorealistic image generation and editing with text-guided diffusion models. arXiv preprint arXiv:2112.10741  (2021)

\bibitem{norman2002two}
Norman, J.: Two visual systems and two theories of perception: An attempt to reconcile the constructivist and ecological approaches. Behavioral and brain sciences  \textbf{25}(1),  73--96 (2002)

\bibitem{otani2023toward}
Otani, M., Togashi, R., Sawai, Y., Ishigami, R., Nakashima, Y., Rahtu, E., Heikkil{\"a}, J., Satoh, S.: Toward verifiable and reproducible human evaluation for text-to-image generation. In: Proceedings of the IEEE/CVF Conference on Computer Vision and Pattern Recognition. pp. 14277--14286 (2023)

\bibitem{pan2023drag}
Pan, X., Tewari, A., Leimk{\"u}hler, T., Liu, L., Meka, A., Theobalt, C.: Drag your gan: Interactive point-based manipulation on the generative image manifold. In: ACM SIGGRAPH 2023 Conference Proceedings. pp. 1--11 (2023)

\bibitem{patashnik2021styleclip}
Patashnik, O., Wu, Z., Shechtman, E., Cohen-Or, D., Lischinski, D.: Styleclip: Text-driven manipulation of stylegan imagery. In: Proceedings of the IEEE/CVF International Conference on Computer Vision. pp. 2085--2094 (2021)

\bibitem{ponomarenko2015image}
Ponomarenko, N., Jin, L., Ieremeiev, O., Lukin, V., Egiazarian, K., Astola, J., Vozel, B., Chehdi, K., Carli, M., Battisti, F., et~al.: Image database tid2013: Peculiarities, results and perspectives. Signal processing: Image communication  \textbf{30},  57--77 (2015)

\bibitem{ponomarenko2009tid2008}
Ponomarenko, N., Lukin, V., Zelensky, A., Egiazarian, K., Carli, M., Battisti, F.: Tid2008-a database for evaluation of full-reference visual quality assessment metrics. Advances of modern radioelectronics  \textbf{10}(4),  30--45 (2009)

\bibitem{prashnani2018pieapp}
Prashnani, E., Cai, H., Mostofi, Y., Sen, P.: Pieapp: Perceptual image-error assessment through pairwise preference. In: Proceedings of the IEEE Conference on Computer Vision and Pattern Recognition. pp. 1808--1817 (2018)

\bibitem{ramesh2022hierarchical}
Ramesh, A., Dhariwal, P., Nichol, A., Chu, C., Chen, M.: Hierarchical text-conditional image generation with clip latents. arXiv preprint arXiv:2204.06125  \textbf{1}(2), ~3 (2022)

\bibitem{de1996naturalness}
de~Ridder, H.: Naturalness and image quality: saturation and lightness variation in color images of natural scenes. Journal of imaging science and technology  \textbf{40}(6),  487--493 (1996)

\bibitem{de1995naturalness}
de~Ridder, H., Blommaert, F.J., Fedorovskaya, E.A.: Naturalness and image quality: chroma and hue variation in color images of natural scenes. In: Human Vision, Visual Processing, and Digital Display VI. vol.~2411, pp. 51--61. SPIE (1995)

\bibitem{rombach2022high}
Rombach, R., Blattmann, A., Lorenz, D., Esser, P., Ommer, B.: High-resolution image synthesis with latent diffusion models. In: Proceedings of the IEEE/CVF conference on computer vision and pattern recognition. pp. 10684--10695 (2022)

\bibitem{russakovsky2015imagenet}
Russakovsky, O., Deng, J., Su, H., Krause, J., Satheesh, S., Ma, S., Huang, Z., Karpathy, A., Khosla, A., Bernstein, M., et~al.: Imagenet large scale visual recognition challenge. International journal of computer vision  \textbf{115},  211--252 (2015)

\bibitem{saharia2022photorealistic}
Saharia, C., Chan, W., Saxena, S., Li, L., Whang, J., Denton, E.L., Ghasemipour, K., Gontijo~Lopes, R., Karagol~Ayan, B., Salimans, T., et~al.: Photorealistic text-to-image diffusion models with deep language understanding. Advances in Neural Information Processing Systems  \textbf{35},  36479--36494 (2022)

\bibitem{salimans2016improved}
Salimans, T., Goodfellow, I., Zaremba, W., Cheung, V., Radford, A., Chen, X.: Improved techniques for training gans. Advances in neural information processing systems  \textbf{29} (2016)

\bibitem{sheikh2005live}
Sheikh, H.: Live image quality assessment database release 2. http://live. ece. utexas. edu/research/quality  (2005)

\bibitem{silberman2018responsible}
Silberman, M.S., Tomlinson, B., LaPlante, R., Ross, J., Irani, L., Zaldivar, A.: Responsible research with crowds: pay crowdworkers at least minimum wage. Communications of the ACM  \textbf{61}(3),  39--41 (2018)

\bibitem{su2020blindly}
Su, S., Yan, Q., Zhu, Y., Zhang, C., Ge, X., Sun, J., Zhang, Y.: Blindly assess image quality in the wild guided by a self-adaptive hyper network. In: Proceedings of the IEEE/CVF Conference on Computer Vision and Pattern Recognition. pp. 3667--3676 (2020)

\bibitem{sun2017mdid}
Sun, W., Zhou, F., Liao, Q.: Mdid: A multiply distorted image database for image quality assessment. Pattern Recognition  \textbf{61},  153--168 (2017)

\bibitem{timofte2017ntire}
Timofte, R., Agustsson, E., Van~Gool, L., Yang, M.H., Zhang, L.: Ntire 2017 challenge on single image super-resolution: Methods and results. In: Proceedings of the IEEE conference on computer vision and pattern recognition workshops. pp. 114--125 (2017)

\bibitem{wang2023unsupervised}
Wang, H., Zhai, D., Liu, X., Jiang, J., Gao, W.: Unsupervised deep exemplar colorization via pyramid dual non-local attention. IEEE Transactions on Image Processing  \textbf{32},  4114--4127 (2023)

\bibitem{wang2023exploring}
Wang, J., Chan, K.C., Loy, C.C.: Exploring clip for assessing the look and feel of images. In: Proceedings of the AAAI Conference on Artificial Intelligence. vol.~37, pp. 2555--2563 (2023)

\bibitem{wang2023aigciqa2023}
Wang, J., Duan, H., Liu, J., Chen, S., Min, X., Zhai, G.: Aigciqa2023: A large-scale image quality assessment database for ai generated images: from the perspectives of quality, authenticity and correspondence. In: CAAI International Conference on Artificial Intelligence. pp. 46--57. Springer (2023)

\bibitem{wang2019detecting}
Wang, S.Y., Wang, O., Owens, A., Zhang, R., Efros, A.A.: Detecting photoshopped faces by scripting photoshop. In: Proceedings of the IEEE/CVF International Conference on Computer Vision. pp. 10072--10081 (2019)

\bibitem{wu2022fast}
Wu, H., Chen, C., Hou, J., Liao, L., Wang, A., Sun, W., Yan, Q., Lin, W.: Fast-vqa: Efficient end-to-end video quality assessment with fragment sampling. In: European Conference on Computer Vision. pp. 538--554. Springer (2022)

\bibitem{wu2023exploring}
Wu, H., Zhang, E., Liao, L., Chen, C., Hou, J., Wang, A., Sun, W., Yan, Q., Lin, W.: Exploring video quality assessment on user generated contents from aesthetic and technical perspectives. In: Proceedings of the IEEE/CVF International Conference on Computer Vision. pp. 20144--20154 (2023)

\bibitem{wu2023q}
Wu, H., Zhang, Z., Zhang, E., Chen, C., Liao, L., Wang, A., Xu, K., Li, C., Hou, J., Zhai, G., et~al.: Q-instruct: Improving low-level visual abilities for multi-modality foundation models. arXiv preprint arXiv:2311.06783  (2023)

\bibitem{wu2023ai}
Wu, J., Gan, W., Chen, Z., Wan, S., Lin, H.: Ai-generated content (aigc): A survey. arXiv preprint arXiv:2304.06632  (2023)

\bibitem{xia2022disentangled}
Xia, M., Hu, W., Wong, T.T., Wang, J.: Disentangled image colorization via global anchors. ACM Transactions on Graphics (TOG)  \textbf{41}(6),  1--13 (2022)

\bibitem{yan2019naturalness}
Yan, B., Bare, B., Tan, W.: Naturalness-aware deep no-reference image quality assessment. IEEE Transactions on Multimedia  \textbf{21}(10),  2603--2615 (2019)

\bibitem{yang2022maniqa}
Yang, S., Wu, T., Shi, S., Lao, S., Gong, Y., Cao, M., Wang, J., Yang, Y.: Maniqa: Multi-dimension attention network for no-reference image quality assessment. In: Proceedings of the IEEE/CVF Conference on Computer Vision and Pattern Recognition. pp. 1191--1200 (2022)

\bibitem{yi2023towards}
Yi, R., Tian, H., Gu, Z., Lai, Y.K., Rosin, P.L.: Towards artistic image aesthetics assessment: a large-scale dataset and a new method. In: Proceedings of the IEEE/CVF Conference on Computer Vision and Pattern Recognition. pp. 22388--22397 (2023)

\bibitem{ying2020patches}
Ying, Z., Niu, H., Gupta, P., Mahajan, D., Ghadiyaram, D., Bovik, A.: From patches to pictures (paq-2-piq): Mapping the perceptual space of picture quality. In: Proceedings of the IEEE/CVF Conference on Computer Vision and Pattern Recognition. pp. 3575--3585 (2020)

\bibitem{yu2019predicting}
Yu, X., Bampis, C.G., Gupta, P., Bovik, A.C.: Predicting the quality of images compressed after distortion in two steps. IEEE Transactions on Image Processing  \textbf{28}(12),  5757--5770 (2019)

\bibitem{zhai2020perceptual}
Zhai, G., Min, X.: Perceptual image quality assessment: a survey. Science China Information Sciences  \textbf{63},  1--52 (2020)

\bibitem{zhan2022bi}
Zhan, F., Yu, Y., Wu, R., Zhang, J., Cui, K., Xiao, A., Lu, S., Miao, C.: Bi-level feature alignment for versatile image translation and manipulation. In: European Conference on Computer Vision. pp. 224--241. Springer (2022)

\bibitem{zhang2023magicbrush}
Zhang, K., Mo, L., Chen, W., Sun, H., Su, Y.: Magicbrush: A manually annotated dataset for instruction-guided image editing. arXiv preprint arXiv:2306.10012  (2023)

\bibitem{zhang2022perceptual}
Zhang, L., Zhou, Y., Barnes, C., Amirghodsi, S., Lin, Z., Shechtman, E., Shi, J.: Perceptual artifacts localization for inpainting. In: European Conference on Computer Vision. pp. 146--164. Springer (2022)

\bibitem{zhang2023adding}
Zhang, L., Agrawala, M.: Adding conditional control to text-to-image diffusion models. arXiv preprint arXiv:2302.05543  (2023)

\bibitem{zhang2023internlm}
Zhang, P., Wang, X.D.B., Cao, Y., Xu, C., Ouyang, L., Zhao, Z., Ding, S., Zhang, S., Duan, H., Yan, H., et~al.: Internlm-xcomposer: A vision-language large model for advanced text-image comprehension and composition. arXiv preprint arXiv:2309.15112  (2023)

\bibitem{zhang2020cross}
Zhang, P., Zhang, B., Chen, D., Yuan, L., Wen, F.: Cross-domain correspondence learning for exemplar-based image translation. In: Proceedings of the IEEE/CVF Conference on Computer Vision and Pattern Recognition. pp. 5143--5153 (2020)

\bibitem{zhang2018unreasonable}
Zhang, R., Isola, P., Efros, A.A., Shechtman, E., Wang, O.: The unreasonable effectiveness of deep features as a perceptual metric. In: Proceedings of the IEEE conference on computer vision and pattern recognition. pp. 586--595 (2018)

\bibitem{zhang2018blind}
Zhang, W., Ma, K., Yan, J., Deng, D., Wang, Z.: Blind image quality assessment using a deep bilinear convolutional neural network. IEEE Transactions on Circuits and Systems for Video Technology  \textbf{30}(1),  36--47 (2018)

\bibitem{zhang2021uncertainty}
Zhang, W., Ma, K., Zhai, G., Yang, X.: Uncertainty-aware blind image quality assessment in the laboratory and wild. IEEE Transactions on Image Processing  \textbf{30},  3474--3486 (2021)

\bibitem{zhang2023blind}
Zhang, W., Zhai, G., Wei, Y., Yang, X., Ma, K.: Blind image quality assessment via vision-language correspondence: A multitask learning perspective. In: Proceedings of the IEEE/CVF Conference on Computer Vision and Pattern Recognition. pp. 14071--14081 (2023)

\bibitem{zhang2023perceptual}
Zhang, Z., Li, C., Sun, W., Liu, X., Min, X., Zhai, G.: A perceptual quality assessment exploration for aigc images. In: 2023 IEEE International Conference on Multimedia and Expo Workshops (ICMEW). pp. 440--445 (2023). \doi{10.1109/ICMEW59549.2023.00082}


\bibitem{zheng2019survey}
Zheng, L., Zhang, Y., Thing, V.L.: A survey on image tampering and its detection in real-world photos. Journal of Visual Communication and Image Representation  \textbf{58},  380--399 (2019)

\bibitem{zheng2022uif}
Zheng, Y., Chen, W., Lin, R., Zhao, T., Le~Callet, P.: Uif: An objective quality assessment for underwater image enhancement. IEEE Transactions on Image Processing  \textbf{31},  5456--5468 (2022)

\bibitem{zhou2017places}
Zhou, B., Lapedriza, A., Khosla, A., Oliva, A., Torralba, A.: Places: A 10 million image database for scene recognition. IEEE transactions on pattern analysis and machine intelligence  \textbf{40}(6),  1452--1464 (2017)

\bibitem{zhou2017scene}
Zhou, B., Zhao, H., Puig, X., Fidler, S., Barriuso, A., Torralba, A.: Scene parsing through ade20k dataset. In: Proceedings of the IEEE conference on computer vision and pattern recognition. pp. 633--641 (2017)

\bibitem{zhu2023genimage}
Zhu, M., Chen, H., Yan, Q., Huang, X., Lin, G., Li, W., Tu, Z., Hu, H., Hu, J., Wang, Y.: Genimage: A million-scale benchmark for detecting ai-generated image. arXiv preprint arXiv:2306.08571  (2023)

\end{thebibliography}

\end{document}